\documentclass{teleai}

\usepackage{amsmath,amsfonts,amssymb}
\usepackage{array,booktabs,graphicx,marvosym,multirow,tabularx}
\usepackage[numbers,sort&compress]{natbib}
\usepackage{url}

\newcommand{\correspondingmark}{\textsuperscript{\Letter}}
\newcolumntype{Y}{>{\centering\arraybackslash}X}
\setcitestyle{square,comma}
\usepackage{booktabs}
\usepackage{array}
\usepackage{makecell}
\usepackage{amssymb} 

\newcolumntype{C}[1]{>{\centering\arraybackslash}p{#1}}

\newcolumntype{L}[1]{>{\raggedright\arraybackslash}p{#1}}

\begin{document}

\title{TourPhysics: Bringing Physics to World Models for Exploration and Manipulation from a Single Image}

\author[1,2]{Xin Zhang\textsuperscript{*}\textsuperscript{$\dagger$}}
\author[2]{Yabo Chen\textsuperscript{*}}
\author[2,3]{Zixuan Duan}
\author[2]{Haibin Huang}
\author[2]{Chi Zhang}
\author[1]{Feng Xu}
\author[2]{Xuelong Li\correspondingmark}

\affiliation[1]{Fudan University}
\affiliation[2]{Institute of Artificial Intelligence (TeleAI), China Telecom}
\affiliation[3]{Nanjing University}


\contribution{\textsuperscript{*}Equal contributions. \textsuperscript{$\dagger$}Work done during an internship at TeleAI. \correspondingmark Corresponding authors.}

\metadata[Keywords]{World models; Physical simulation; Video generation; Single-image scene reconstruction; Persistent memory; Camera control; Physical interaction}

\abstract{Interactive visual world models must distinguish observation from physical intervention. Camera motion reveals new surfaces, whereas intervention changes object motion, contact, and deformation. Current video world models are largely driven by appearance priors and often lose physical or spatial consistency over long horizons. We present TourPhysics, an online framework initialized from a single image and a declarative physical configuration. TourPhysics extends PhysOmni, our ACM Multimedia 2026 work, from finite physics-grounded video synthesis to persistent exploration and manipulation. TourPhysics combines deterministic simulation with video generation while assigning separate roles to simulator state, geometric evidence, generator controls, and appearance memory. For each action, the simulator computes a finite physical and camera trajectory before the corresponding observation is generated. Accepted observations publish the terminal state and update the appearance memory and subsequent generator controls, while the committed state and simulator geometry remain fixed throughout synthesis and retry. We further separate the simulator geometry used for projection and visibility from the relative depth used to condition the generator. A reference-anchored memory retrieves accepted static appearance through geometric cross-view correspondence and incorporates it through a bounded residual that reverts to the native path when no valid correspondence exists. On simulator-defined camera tours and object manipulations, TourPhysics follows prescribed camera and object trajectories more closely than the evaluated baselines, preserves the input scene, and reduces appearance drift during long-horizon revisits.}

\maketitle

\begin{figure}[!t]
    \centering
    \includegraphics[pagebox=mediabox,width=\linewidth,trim=19pt 37pt 19pt 24pt,clip]{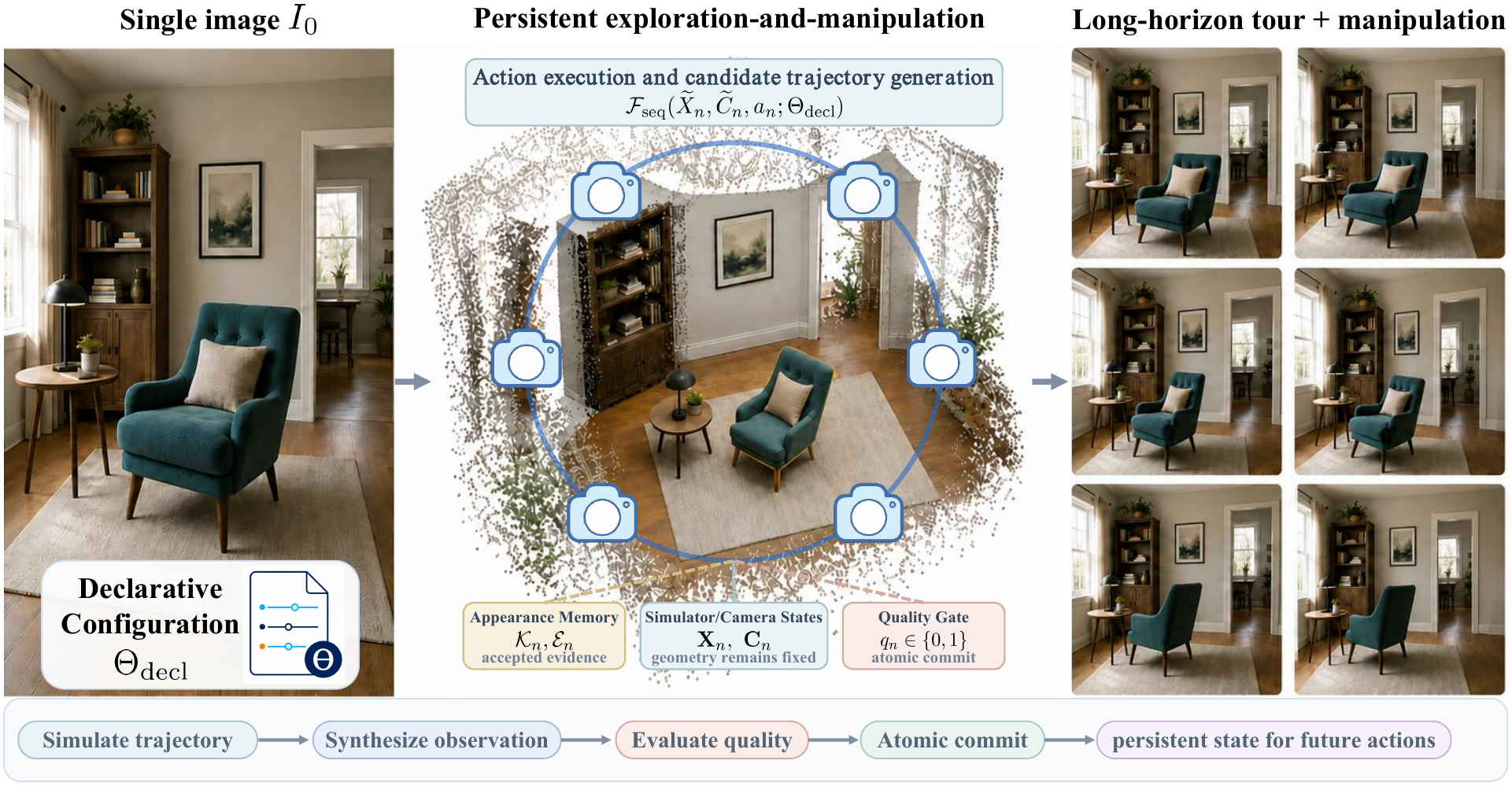}
    \caption{Given a single image \(I_0\) and a declarative configuration \(\Theta_{\mathrm{decl}}\), TourPhysics maintains a persistent scene model for camera exploration and object manipulation. For each action, a physical transition computes a fixed physical and camera trajectory; the system then synthesizes the corresponding observation, applies a quality gate, and atomically commits the simulator state together with the accepted appearance evidence. The simulator state and geometric record remain fixed during synthesis, and the committed state supports the long-horizon tour and manipulation shown on the right.}
    \label{fig:teaser}
\end{figure}

\section{Introduction}\label{sec:intro}

An interactive world model must distinguish observation from intervention. Moving a camera reveals previously unseen surfaces, whereas applying a force changes object positions, contacts, and deformations. Both operations produce new images, but only intervention changes the physical state; observation instead expands the knowledge that the system holds of the same scene.

Recent video generators produce convincing camera motion and object dynamics, and interactive world models increasingly accept viewpoint and action controls~\cite{wan2025open,kong2024hunyuanvideo,openai_sora2_2025,bruce2024genie,deepmind2024genie2,decart2024oasis,robbyant2026lingbot}. Most of these systems rely on learned appearance priors, so plausible video does not guarantee a persistent or physically faithful world. A scene may change when revisited, temporal caches may carry hallucinated details forward, and learned motion may ignore a prescribed interaction. These failures arise when generated appearance is allowed to stand in for geometric evidence or physical state.

Physics-aware generation methods add trajectories, forces, material controls, or simulator outputs to video synthesis~\cite{zhang2024physdreamer,liu2024physgen,wang2025physctrl}. Geometry-aware methods use camera poses, depth, or reconstructed scene evidence to improve viewpoint control and cross-view consistency~\cite{yu2025viewcrafter,seo2024genwarp}. Combining these capabilities requires explicit ownership of state. Our work builds on PhysOmni~\cite{zhang2026physomni} and extends its single-image, physics-grounded, multi-object scene generation and real-time interaction setting to persistent exploration and manipulation. While PhysOmni focuses on generating and interacting with a physics-grounded scene from a single image, TourPhysics maintains an explicit world state across multiple action--observation cycles. Each cycle consists of a simulator-defined physical and camera trajectory followed by observation synthesis over that fixed trajectory segment. A quality-gated commit then either publishes the terminal state and the accepted appearance evidence or leaves the persistent world unchanged. In this way, TourPhysics turns physics-grounded generation into an online, stateful world-model loop: the simulator governs object evolution, simulator-consistent geometry governs projection and visibility, and the generator supplies appearance without altering the physical scene model.

Single-image initialization makes this separation especially important. An image constrains visible appearance but leaves occluded geometry, scale, material parameters, and future dynamics underdetermined. TourPhysics combines the image with a user-provided declarative configuration to build one persistent, controllable physical hypothesis. The configuration specifies controllable objects, material families, boundary conditions, interventions, and camera actions.

TourPhysics processes one committed observation window at a time. For each action, it computes a finite physical and camera trajectory and then generates an observation of that fixed segment. A quality gate either commits the terminal state and the appearance evidence together or leaves the persistent state unchanged. Generated observations may extend appearance knowledge, but they cannot modify the committed dynamics or the simulator geometry.

Figure~\ref{fig:teaser} summarizes this workflow, from a single image to an interactive world, and illustrates the resulting long-horizon camera exploration and object manipulation.

TourPhysics assigns different roles to two depth representations. Simulator-consistent depth governs projection, visibility, correspondence, and trajectory evaluation, while relative depth conditions the video generator and may be completed from accepted observations. For long-horizon consistency, the system retains the original input image as a permanent reference and retrieves the accepted appearance through geometric cross-view correspondence. Historical evidence enters attention as a bounded residual; when no valid correspondence is available, the model follows its native reference-conditioned path.

We evaluate TourPhysics on simulator-defined camera tours and object manipulations involving rigid, cloth, elastic, granular, and fluid proxies. The evaluation measures alignment with prescribed camera and object trajectories, event timing and deformation, input-scene similarity, and long-horizon appearance stability.

Our contributions are twofold:
\begin{itemize}
    \item We formulate single-image tour-and-manipulate generation as coupled physical and epistemic transitions and present TourPhysics as an online world model with explicit state ownership. The simulator fixes the physical and camera trajectory before observation synthesis, while a quality-gated commit publishes the accepted state and visual evidence atomically.

    \item We introduce two complementary consistency mechanisms: simulator-consistent geometry is kept separate from generator-facing depth, and a reference-anchored memory retrieves accepted static appearance through geometry-routed cross-view correspondence. Historical features enter as a confidence-bounded residual and fall back exactly to the native reference-conditioned path when correspondence is invalid.

\end{itemize}

\section{Related Work}\label{sec:related}

\subsection{Video World Models and Interactive Environments}
Diffusion models supply strong appearance and motion priors for video generation~\cite{croitoru2023diffusion,blattmann2023svd,wan2025open,kong2024hunyuanvideo,openai_sora2_2025,tpami_video_prediction_review,tpami_vit_survey,huang2024domainfusion}, while recent analyses of foundation models summarize the scaling and conditioning trends underlying these systems~\cite{tpami_foundation_models}. Complementary AI Flow studies consider system-level AI service organization~\cite{an2025aiflowperspectivesscenarios,shao2025aiflownetworkedge}. Studies of temporal prediction and interpolation further reveal the need to preserve motion and appearance across adjacent frames~\cite{tpami_memcnet,tpami_video_inpainting}, whereas interactive world models synthesize observations conditioned on learned actions or compact state representations~\cite{bruce2024genie,yang2023unisim,decart2024oasis,che2024gamegenx,valevski2024gamengen,chen2025teleworlddynamicmultimodalsynthesis}. Playable Environments represents a scene from a single image through controllable camera poses and learned object actions~\cite{menapace2022playable}, whereas Dexterous World Models conditions video diffusion on a static 3D rendering together with an egocentric hand trajectory~\cite{kim2026dexterous}. Although these methods support action-controlled video synthesis, they generally do not distinguish observation from intervention within their state updates. In TourPhysics, only the physical branch updates the simulator state, whereas camera-induced observations update appearance knowledge only after they commit.

\subsection{Geometry-Aware Video and Persistent Scene Representations}
Camera-controlled video generation relies on camera pose, depth, optical flow, or reconstructed 3D evidence to improve viewpoint control and cross-view consistency~\cite{cai2024generativerendering,yu2025viewcrafter,he2025cameractrlii,zhang2026symphomotion,tpami_novel_views,tpami_surRF,tpami_refnerf,chen2026full4dgeneratingfullscope4d}. Persistent approaches maintain point clouds or explicit 3D/4D representations across generated views~\cite{ren2025gen3c,zhang2026worldstereo,chen2025v3d,wang2025video4dgen,tpami_pointcloud_survey,tpami_nonrigid_registration,tpami_dense3d}. Single-image methods estimate camera parameters and dense geometry, or construct object-centric and scene-level representations~\cite{park2024singleview,chen2023scenedreamer,wang2025vggt,wang2025moge,wu2025diorama,yao2025cast,xiang2025trellis,wu2025amodal3r,tpami_geonetpp,tpami_depth_completion,tpami_epc,tpami_pixel2meshpp,chen2024cascadezero123imagehighlyconsistent,chen2024liftimage3dliftingsingleimage,wen2025metricsolverslidinganchoredmetric}. CUT3R is closely related, as it maintains a shared metric pointmap as observations arrive~\cite{wang2025cut3r}. Nevertheless, monocular scale, topology, and occlusion remain ambiguous~\cite{arampatzakis2024monocular}. TourPhysics uses visible geometry together with a declarative configuration to initialize a controllable physical hypothesis, reserves simulator depth for projection and visibility, and employs separate depth to condition the generator.

\subsection{Physics-Grounded and Simulator-Guided Generation}
Physics-aware generation incorporates learned dynamics, trajectories, forces, material controls, or explicit simulation into image and video synthesis~\cite{ma2026miramo,li2024generativedynamics,shen2026phantom,zhang2024physdreamer,liu2024physgen,tan2024physmotion,wang2025physctrl,gillman2025forceprompting,huang2025zero}. Related TPAMI work on physics-based generative restoration couples image priors with explicit degradation constraints~\cite{tpami_physics_gan}, while masked operators such as partial convolution restrict updates to the valid support during inpainting and synthesis~\cite{tpami_partialconv}. PhysGen3D constructs an amodal, camera-centric world from user-specified initial conditions and material properties~\cite{chen2025physgen3d}. Physical Simulator In-the-Loop Video Generation reconstructs a generated template, simulates foreground trajectories, and uses simulator correspondences to guide synthesis~\cite{foo2026physicalsim}. TourPhysics instead computes each declared action with a rigid-body, PBD~\cite{muller2007pbd}, or MPM~\cite{jiang2016material} model before the video model observes the resulting fixed state sequence.

\subsection{Long-Video Memory and Reference Conditioning}
Long-video methods extend temporal support through sliding windows, recurrent contexts, or bounded attention caches. Related autoregressive work improves long-horizon planning, test-time correction, and attention efficiency~\cite{xiang2025macrofrommicroplanninghighqualityparallelized,xiang2026pathwisetesttimecorrectionautoregressive,wang2026directingworldfastautoregressive,xiang2025makeefficientdynamicsparse}. Reference-conditioned and memory-based approaches transfer texture or recover long-range context through correspondence mechanisms, dynamic memory, and cross-attention~\cite{jiang2022c2matching,hu2025drsa,huang2026cineweaver,tpami_vos_memory}. Established surveys on visual tracking provide complementary correspondence and temporal-association baselines for maintaining identity under motion and occlusion~\cite{tpami_tracking_survey,tpami_tracking_experimental}. LongVie supports multimodal long-video generation~\cite{gao2025longviemultimodalguidedcontrollableultralong}, whereas Deep Forcing introduces a training-free cache rule to retain early context under finite attention capacity~\cite{yi2025deepforcing}. Although these mechanisms improve temporal continuity, they do not explicitly associate historical appearance with persistent geometric surfaces or committed simulator states. TourPhysics augments a finite-window video diffusion model with geometry-routed memory, in which an accepted historical key/value page is selected through cross-view visibility and incorporated as a confidence-bounded residual around the permanent real-image reference.

\begin{figure}[!t]
    \centering
    \includegraphics[pagebox=mediabox,width=\linewidth,trim=3pt 136pt 2pt 3pt,clip]{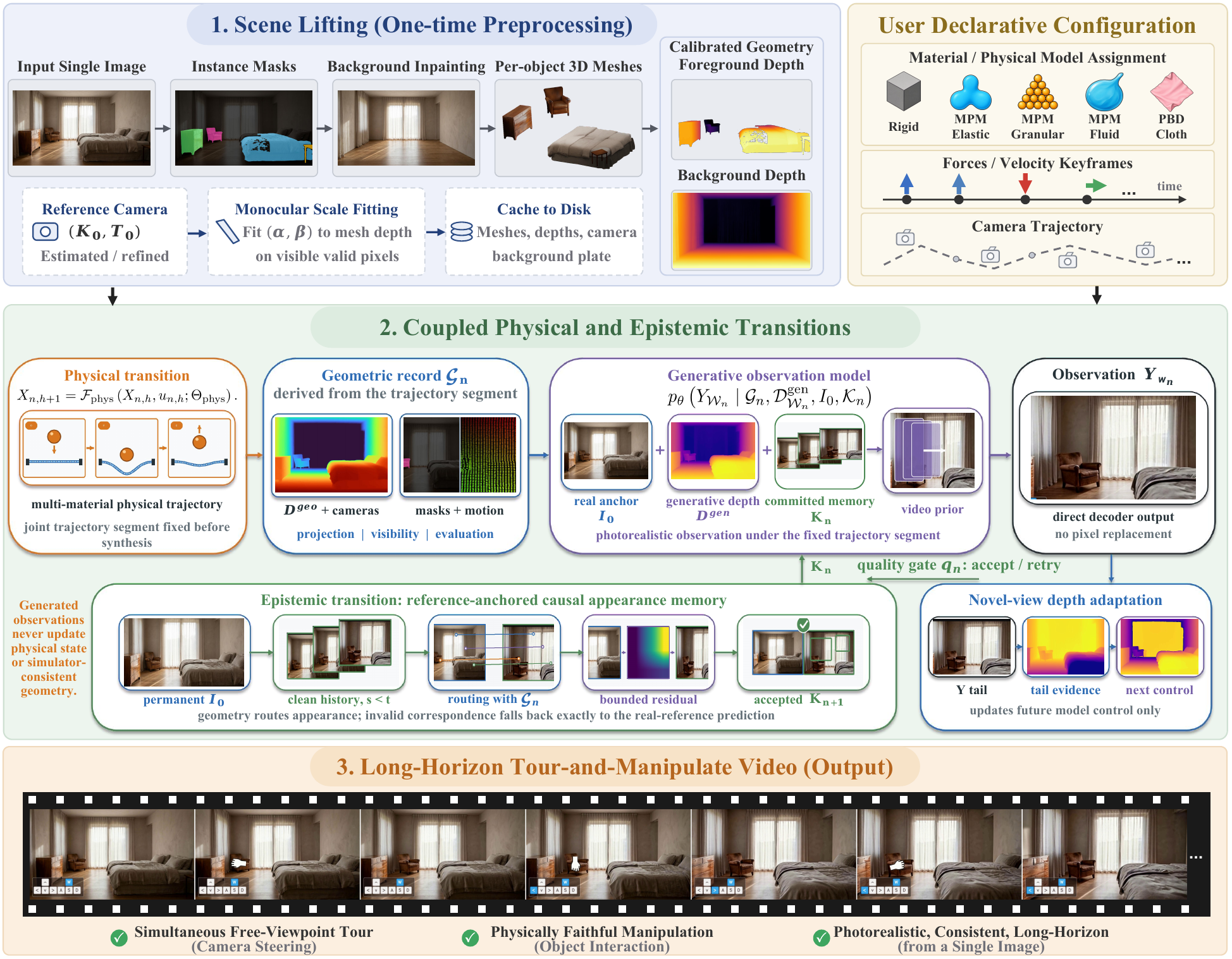}
    \caption{Overview of TourPhysics. (1) \emph{Scene Lifting (One-time Preprocessing)} combines the input image with a declarative user configuration, including material models, force and velocity keyframes, and camera commands, to construct a persistent scene hypothesis comprising instance masks, a background plate, per-object 3D meshes, calibrated foreground and background geometry, a reference camera, and cached simulator assets. (2) \emph{Coupled Physical and Epistemic Transitions} advances one action window. The orange \emph{Physical transition} block fixes the multi-material physical and camera segment before synthesis; the read-only geometric record \(\mathcal G_n\) supplies simulator-consistent depth, cameras, masks, and motion; and the observation model combines these controls with generator-facing depth \(\mathcal D_{\mathcal W_n}^{\mathrm{gen}}\), the real-image anchor \(I_0\), and committed appearance memory \(\mathcal K_n\) to produce \(Y_{\mathcal W_n}\). A quality gate publishes the accepted state and appearance atomically; only after acceptance do reference-anchored memory and novel-view depth adaptation affect subsequent appearance controls. (3) \emph{Long-Horizon Tour-and-Manipulate Video (Output)} chains the accepted windows into a persistent sequence that supports free-viewpoint exploration, physically faithful manipulation, and photorealistic long-horizon consistency, while the generated RGB never replaces simulator renderings or writes back to the physical state or simulator geometry.}
    \label{fig:framework}
\end{figure}

\section{Problem Formulation}\label{sec:problem}

Given a single RGB image \(I_0\) and a declarative configuration \(\Theta_{\mathrm{decl}}\), we aim to construct an online world model that supports both camera exploration and physical manipulation. The desired
output is a long-horizon sequence of observations that follows the declared physical--camera trajectory while maintaining appearance consistency through committed state transitions across accepted windows. The formulation is
organized into three parts: Figure~\ref{fig:framework}(1),
\emph{Scene Lifting (One-time Preprocessing)}, which constructs the persistent scene hypothesis and reference; Figure~\ref{fig:framework}(2), \emph{Coupled Physical and Epistemic Transitions}, which advances a single
committed window; and Figure~\ref{fig:framework}(3),
\emph{Long-Horizon Tour-and-Manipulate Video (Output)}, which produces the resulting observation sequence.

This separation makes the ownership of state explicit: the physical branch
determines the simulator state and geometry, the observation branch
synthesizes observations for the resulting fixed segment, and the epistemic
branch publishes only accepted appearance evidence. The resulting causal
order is illustrated in the bottom protocol of Figure~\ref{fig:teaser}:
simulate the trajectory, synthesize the corresponding observation, evaluate
the quality gate, and atomically commit the accepted result for subsequent
actions. In particular, the committed simulator and camera states serve as
the starting point for the next action, while the accepted appearance memory
and tail evidence provide conditioning context for subsequent generation.

\subsection{Scene Lifting and Persistent Reference}
\label{sec:problem_scene_lifting}

The input is given by
\[
\mathcal I_{\mathrm{init}}
=
(I_0,\Theta_{\mathrm{decl}}),
\]
where \(\Theta_{\mathrm{decl}}\) specifies the controllable objects, material
and physical-model assignments, boundary conditions, interventions, and
camera trajectory controls. The one-time scene-lifting stage constructs an
internally consistent scene hypothesis from this underdetermined input,
estimates a reference camera and calibrated geometry, and preserves \(I_0\),
together with any cached background plate, as permanent reference
information. These reference records and the declaration remain fixed
throughout online execution and are read by every subsequent operator.

At the boundary preceding the execution of action \(a_n=(c_n,u_n)\), the
committed state is given by
\[
    \mathcal S_n=(X_n,C_n,\mathcal K_n,\mathcal E_n).
\]
The index \(n\) counts the accepted windows. A rejected attempt is treated
as a retry of the same action, indexed by \(r\), and leaves both \(n\) and
\(\mathcal S_n\) unchanged; for notational simplicity, we suppress \(r\)
below.

\subsection{Coupled Physical and Epistemic Transitions}
\label{sec:problem_transitions}

\paragraph{Physical transition and geometric record.}

At a committed boundary, the camera command \(c_n\) and the optional
intervention \(u_n\) are resolved into a framewise physical--camera
trajectory before any pixels are sampled. The simulator and camera are
advanced from working copies of the preceding committed checkpoint,
\[
(\widetilde X_n,\widetilde C_n)
=
\operatorname{clone}(X_n,C_n),
\]
so that a candidate attempt produces no persistent side effects.

The physical and camera trajectories, together with the associated geometric
record, are then produced by
\begin{equation}
    \left(
    \mathbf X_n,\mathbf C_n,\mathcal G_n
    \right)
    =
    \mathcal F_{\mathrm{seq}}
    \left(
    \widetilde X_n,\widetilde C_n,a_n;
    \Theta_{\mathrm{decl}}
    \right).
    \label{eq:online_sequence}
\end{equation}

Here,
\[
\begin{aligned}
\mathbf X_n
&=
\left(
X_n^{(0)},X_n^{(1)},\ldots,X_n^{(T_n)}
\right),\\
\mathbf C_n
&=
\left(
C_n^{(0)},C_n^{(1)},\ldots,C_n^{(T_n)}
\right)
\end{aligned}
\]
denote the framewise physical and camera state sequences, respectively. The
index \(t=0\) denotes the committed boundary state, while
\(t=1,\ldots,T_n\) index the newly generated states within the candidate
window:
\[
X_n^{(0)}=\widetilde X_n,
\qquad
C_n^{(0)}=\widetilde C_n.
\]
The terminal working states are given by
\[
X_n^+=X_n^{(T_n)},
\qquad
C_n^+=C_n^{(T_n)}.
\]

The candidate record \(\mathcal G_n\) contains simulator-consistent
framewise depth, camera parameters, object masks, motion, visibility,
correspondence, and validity information derived from
\((\mathbf X_n,\mathbf C_n)\).

\paragraph{Generative observation model.}

Once the candidate window \(\mathcal W_n\) and its geometric record have
been fixed, the observation model takes as input the geometric record, the
generator-facing depth \(\mathcal D_{\mathcal W_n}^{\mathrm{gen}}\), the
permanent real-image anchor \(I_0\), and the committed appearance memory
\(\mathcal K_n\), and produces a candidate observation.

Prior to sampling, the generator-facing depth is constructed from the fixed
geometric record and the accepted tail evidence:
\[
\mathcal D_{\mathcal W_n}^{\mathrm{gen}}
=
\mathcal F_{\mathrm{ctrl}}
\left(
\mathcal G_n,\mathcal E_n
\right).
\]
Once constructed, the full framewise control tensor remains fixed for the
duration of the current candidate window and is not recomputed during
observation sampling:
\begin{equation}
    Y_{\mathcal W_n}
    \sim
    p_\theta\!\left(
    Y_{\mathcal W_n}
    \mid
    \mathcal G_n,
    \mathcal D_{\mathcal W_n}^{\mathrm{gen}},
    I_0,
    \mathcal K_n
    \right).
    \label{eq:problem_observation_model}
\end{equation}

Because \(\mathcal G_n\) encodes the simulator-consistent framewise
geometric and camera records derived from the physical and camera
trajectories, the trajectory information required by the observation model
is supplied implicitly through \(\mathcal G_n\). Although
\(\mathcal D_{\mathcal W_n}^{\mathrm{gen}}\) is deterministically derived
from \(\mathcal G_n\) and \(\mathcal E_n\), it is supplied explicitly
because it constitutes the representation consumed by the generator. The
conditional form above therefore specifies the generative observation
model.

The real-image anchor \(I_0\) is immutable and is used solely as a
conditioning reference. Regions that are not observed in \(I_0\), including
regions exposed by the current camera trajectory, may be synthesized using
the fixed geometric record, the accepted tail evidence, and the committed
appearance memory. Simulator-derived signals may provide conditioning,
geometric, and evaluation information, but simulator RGB is not used as a
direct substitute for generated pixels. Consequently, the observation branch
may extend the appearance knowledge through accepted generated
observations, but it cannot modify \(X_n\), \(C_n\), or the
simulator-consistent geometry that produced the current candidate window.

After sampling, the candidate observation is used to construct the staged
appearance memory and staged tail evidence:
\begin{equation}
    \left(
    \widehat{\mathcal K}_{n+1},
    \widehat{\mathcal E}_{n+1}
    \right)
    =
    \mathcal U_{\mathrm{stage}}
    \left(
    \mathcal K_n,\mathcal E_n,
    Y_{\mathcal W_n},\mathcal G_n
    \right).
    \label{eq:staged_update}
\end{equation}
These staged quantities are candidate-local and remain private until the
candidate window passes the quality gate. If the candidate is rejected,
they are discarded together with the working trajectory, and the committed
appearance memory and accepted tail evidence remain unchanged.

\paragraph{Epistemic transition and quality-gated commit.}

The candidate appearance memory and tail evidence remain private while the
current window is being evaluated. A binary quality decision is computed as
\begin{equation}
    q_n
    =
    Q\!\left(
    Y_{\mathcal W_n},
    \mathcal G_n,
    \mathcal D_{\mathcal W_n}^{\mathrm{gen}},
    \mathcal K_n,
    \mathcal E_n
    \right)
    \in\{0,1\},
    \label{eq:quality_gate}
\end{equation}
which controls the atomic state transition. Here, \(\widehat{\mathcal
K}_{n+1}\) and \(\widehat{\mathcal E}_{n+1}\) denote the staged memory and
tail-evidence updates proposed by the current candidate observation:
\begin{equation}
\begin{aligned}
q_n=1:\quad&
\mathcal S_{n+1}
=
\left(
X_n^+,
C_n^+,
\widehat{\mathcal K}_{n+1},
\widehat{\mathcal E}_{n+1}
\right),\\
q_n=0:\quad&
\mathcal S_n
\text{ remains unchanged}.
\end{aligned}
\label{eq:knowledge_transition}
\end{equation}

An accepted window atomically publishes the terminal simulator and camera
states together with the staged visual evidence. A rejected attempt
discards its working copy and leaves the committed checkpoint \(\mathcal
S_n\) unchanged. A retry retains the same action, trajectory, geometric
record, generator-facing controls, and committed memory, while allowing
only a new sampling seed. The epistemic branch therefore updates
\(\mathcal K_{n+1}\) and the evidence used for future controls only after
acceptance.

\subsection{Long-Horizon Tour-and-Manipulate Video (Output)}
\label{sec:problem_output}

The final output is the sequence of committed observations. To avoid
duplicated boundary frames, the first frame of every subsequent window is
removed prior to concatenation:
\begin{equation}
    \mathsf{Video}
    =
    \operatorname{concat}\!\left(
    Y_{\mathcal W_0},
    Y_{\mathcal W_1}[1{:}],
    \ldots,
    Y_{\mathcal W_{N-1}}[1{:}]
    \right).
\end{equation}
Here, \(N\) denotes the total number of accepted windows; rejected attempts
do not contribute output frames.

The resulting video must support both free-viewpoint camera exploration and physically faithful object manipulation while remaining photorealistic and consistent over long horizons. These requirements are evaluated with
respect to the declared trajectory and the persistent committed state,
rather than through unconstrained visual continuation. The implementation
of each operator is described in Section~\ref{sec:method}.

\section{Methods}\label{sec:method}

\subsection{Overview}

The method follows the causal order of
\emph{scene lifting}, \emph{physical/camera transition},
\emph{geometric record}, \emph{observation synthesis}, \emph{staged update},
\emph{quality gate}, and \emph{atomic commit} (Fig.~\ref{fig:framework}).
Scene lifting builds one editable hypothesis from \(I_0\) and
\(\Theta_{\mathrm{decl}}\); at each accepted boundary, the simulator and camera
controller clone the committed state and produce the complete trajectory and
the read-only record \(\mathcal G_n\). The physical segment and its geometry
are fixed before RGB sampling, and a retry reuses them.

The observation model then consumes \(\mathcal G_n\), a separate
generator-facing depth control, the permanent image anchor, and the committed
appearance context. Memory and tail-depth updates remain candidate-local until
the quality gate; a passing window commits the terminal simulator/camera state
and the staged evidence together, while a failing window rolls back the
candidate. Generated RGB can therefore extend appearance knowledge without
modifying physical state or simulator geometry.

\subsection{Scene Lifting and Persistent Reference}\label{sec:method_lifting}

Scene lifting is performed once, prior to the first interaction. Given the
input image \(I_0\) and the frozen declaration \(\Theta_{\mathrm{decl}}\), the
initializer segments the selected objects, lifts their masks to explicit
meshes, removes the foreground to form a cached background plate \(B_0\), and
aligns the meshes to a shared gravity-aware scene frame. A reference camera and
an affine depth calibration are then fitted from the visible image evidence.
These operations define a single controllable physical hypothesis; they are
not repeated as online state corrections.

\paragraph{Object decomposition and background.}
For each object selected by the user, SAM~3 and SAM~3D Objects produce an
instance mask, a local triangular mesh, and an object-to-camera similarity
transform. These explicit surfaces, rather than the RGB image, serve as the
contact and collision interfaces used by the simulator. The selected
foreground union is removed cumulatively with LaMa
inpainting~\cite{tpami_partialconv,tpami_video_inpainting} to form the cached
plate \(B_0\), while the unmodified \(I_0\) is retained as the
permanent appearance anchor. The plate is used only when a camera move exposes
background support; it is not treated as a new physical object.

The independently lifted meshes are placed in a shared gravity-aware frame. We
pool the lowest \(5\%\) of vertices as provisional contact anchors (relaxed to
\(10\%\) when an object has too few anchors), fit a robust ground plane, and
apply one bounded rigid alignment. Small residual overlaps are corrected along
the minimum-overlap bounding-box axis. These corrections belong to the initial
hypothesis and are never applied online.

The declaration may be authored directly or proposed by the optional
vision-language initializer. In either case, the material and force-field
entries are checked against the supported families, completed with defaults,
exposed for editing, and frozen before the first trajectory simulation. We use
the fixed projections \(\Theta_{\mathrm{phys}}=\Pi_{\mathrm{phys}}(\Theta_{\mathrm{decl}})\)
and \(\Theta_{\mathrm{cam}}=\Pi_{\mathrm{cam}}(\Theta_{\mathrm{decl}})\). An
online action selects a camera command and an intervention for the next finite
window; it does not replace the declaration.

The reference camera is refined by aligning the rendered union silhouette with
the selected masks. For valid pixels, the monocular depth is placed into the
simulator metric convention with a single affine fit,
\begin{equation}
D_0^{\mathrm{geo}}(p)=\alpha\widetilde D_0(p)+\beta ,
\label{eq:method_depth_alignment}
\end{equation}
where invalid and non-finite samples are excluded. Depth/normal estimation and
sparse-depth completion studies motivate treating this fit as a calibrated
geometric control rather than a unique reconstruction~\cite{tpami_geonetpp,tpami_depth_completion};
single-image novel-view work likewise highlights the disocclusion ambiguity
that remains after calibration~\cite{tpami_novel_views}. This calibration fixes
the scale used by later projection and visibility tests.

The immutable reference record is
\begin{equation}
\mathcal R=(I_0,B_0,D_0^{\mathrm{geo}},V_0,Q_0,L_0,K_0,T_0).
\label{eq:method_reference_record}
\end{equation}
Together with the initialized physical and camera states, it defines
\(\mathcal S_0=(X_0,C_0,\mathcal K_0,\mathcal E_0)\), where \(\mathcal K_0\)
contains no generated pages and \(\mathcal E_0\) contains no accepted tail
evidence. The reference image, its calibrated geometry, and the cached
background plate remain read-only throughout online execution. Camera
intrinsics \(K_t\) and the appearance memory \(\mathcal K_n\) are distinct
quantities.

\subsection{Physical and Camera Transition}\label{sec:method_transition}

At the boundary of accepted window \(n\), the committed state is
\(\mathcal S_n=(X_n,C_n,\mathcal K_n,\mathcal E_n)\). For action
\(a_n=(c_n,u_n)\), the simulator and camera controller first clone the
committed physical and camera states. The fixed candidate trajectory and its
geometric record are then produced by the sequence operator already defined in
Eq.~\eqref{eq:online_sequence}:
\(\left(\mathbf X_n,\mathbf C_n,\mathcal G_n\right)
=\mathcal F_{\mathrm{seq}}(\widetilde X_n,\widetilde C_n,a_n;
\Theta_{\mathrm{decl}})\). The terminal candidates \(X_n^+\) and \(C_n^+\)
are the endpoints of this working segment.

Within this fixed segment, each simulator substep applies the declared
physical map to the working state, while the camera controller advances in
parallel from \(c_n\). The first substep starts at the cloned committed
boundary, and the terminal candidate is taken only at the declared sampled
endpoint. Keeping this boundary convention explicit ensures that a retry
reproduces the same physical trajectory instead of inadvertently advancing it
twice.

\begin{figure}[t]
    \centering
    \includegraphics[pagebox=mediabox,width=0.75\linewidth,trim=2pt 31pt 363pt 42pt,clip]{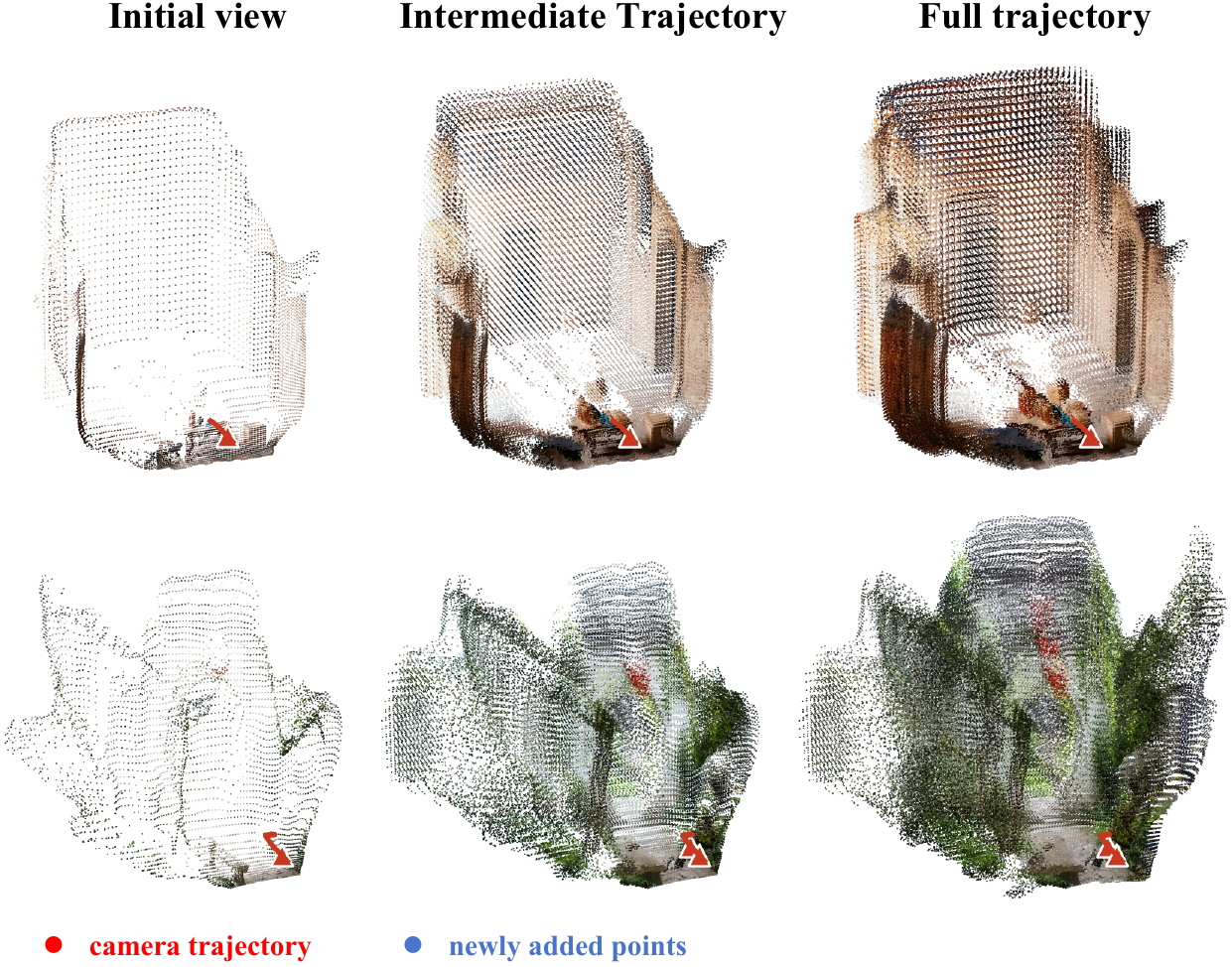}
    \caption{Point-cloud visualization of the camera update. Columns show the
    initial, intermediate, and full-trajectory views; red arrows denote camera
    motion and blue points denote newly observed support.}
    \label{fig:orbit_pointcloud}
\end{figure}

Camera coverage along the fixed update is illustrated in
Fig.~\ref{fig:orbit_pointcloud}; the point cloud is diagnostic and does not
modify simulator geometry or physical state.

\begin{figure}[t]
    \centering
    \includegraphics[width=0.75\linewidth]{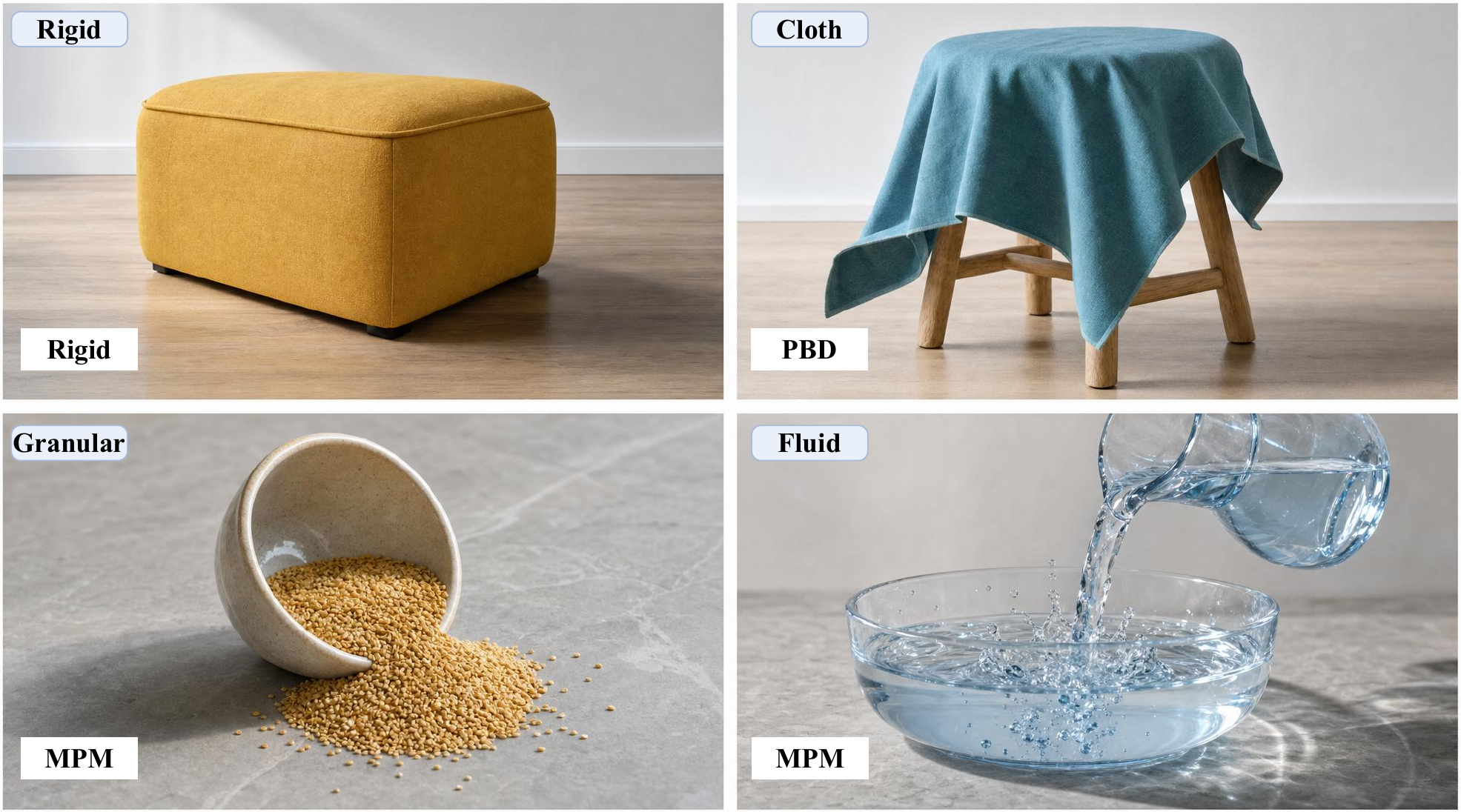}
    \caption{Material families and solver routing used by the physical backend.
    The declarative configuration selects the material model and its parameters;
    the image is a schematic taxonomy, not a generated-video result.}
    \label{fig:material_taxonomy}
\end{figure}

The declaration routes rigid and articulated proxies to RBD, continuum and
volumetric materials to MPM, and thin or constraint-dominated systems to PBD.
Figure~\ref{fig:material_taxonomy} summarizes these material families and their
solver routing.
Gravity, scheduled force fields, and ongoing dynamics continue even when
\(u_n=\varnothing\), so a camera-only action can still advance the physical
state. RBD--MPM and RBD--PBD contacts are coupled within the same scene step.
Unless overridden by the declaration, the solver step is \(4\,\mathrm{ms}\),
with 10 substeps for rigid-only scenes and 16 for particle scenes; gravity is
\((0,0,-9.8)\). Constitutive equations and material-specific parameters are
reported in the supplementary appendix, Sec.~A.2 (``Declarative Configuration
and Physical Simulation''), as they do not affect the observation transaction.

At each frame, the renderer exports the simulator-consistent metric depth,
valid-visible support \(V_{n,t}\), confidence \(Q_{n,t}\), object labels
\(L_{n,t}\), the foreground mask \(M^{\mathrm{mask}}_{n,t}\), motion
\(U^{\mathrm{mot}}_{n,t}\), camera intrinsics, and world-to-camera extrinsics.
These fields form \(\mathcal G_n\) and serve as read-only conditioning and
evaluation signals; they never replace decoded RGB. The reported profile uses
81 decoded frames with one shared conditioning boundary and 80 new frames. The
exact simulator-to-video index convention is given in the supplementary
appendix, Sec.~A.4 (``Observation Generation and Transactional State''); the
endpoint convention introduced there also defines \(X_n^+\) and \(C_n^+\). The complete segment is held
fixed during both sampling and retry.

For clarity, the record used by the generator and by the later memory tests is
\begin{equation}
\mathcal G_n =
\left\{
\begin{gathered}
D_{n,t}^{\mathrm{geo}},\; V_{n,t},\; Q_{n,t},\; L_{n,t},\\
M_{n,t}^{\mathrm{mask}},\; U_{n,t}^{\mathrm{mot}},\;
K_{n,t},\; T_{n,t}
\end{gathered}
\right\}_{t\in\mathcal W_n}.
\label{eq:method_geometry_record}
\end{equation}

It is rendered once from the fixed trajectory and is never updated from
generated frames. With \(f_{\mathrm{sim}}=60\) and \(f_{\mathrm{gen}}=16\), the
center-of-bin mapping used in the evaluated profile is
\begin{equation}
k_n(j)=o_n+\left\lfloor
\frac{(2j+1)f_{\mathrm{sim}}}{2f_{\mathrm{gen}}}
\right\rfloor,\qquad o_{n+1}=o_n+300 .
\label{eq:method_timeline}
\end{equation}
Thus \(k_n(80)=k_{n+1}(0)\) is the shared boundary; \(X_n^+\) and \(C_n^+\)
are defined at this sampled endpoint rather than at the raw origin
\(o_n+300\). This convention removes the off-by-one ambiguity between
simulator substeps and decoded frames.

\subsection{Geometric Record and Generator Controls}\label{sec:method_geometry}

The method keeps the simulator geometry separate from the depth
representation consumed by the video generator. \(D_{n,t}^{\mathrm{geo}}\) is
the metric camera-space \(z\)-depth in the calibrated scene units and is
written only by scene initialization and simulator rendering. It controls
projection, visibility, correspondence, and trajectory evaluation. In
contrast, \(\mathcal D_{\mathcal W_n}^{\mathrm{base}}
=\mathcal F_{\mathrm{base}}(\mathcal G_n;\mathcal R)\) is a normalized
relative-depth condition for the video model. This explicit split follows the
broader practice of keeping metric reconstruction and appearance-oriented
representation separate in point-cloud, dense-reconstruction, and radiance-field
systems~\cite{tpami_pointcloud_survey,tpami_dense3d,tpami_surRF,tpami_refnerf}.

For \(t\in\mathcal W_n\), we retain the three quantities separately:
\begin{equation}
D_{n,t}^{\mathrm{geo}},\qquad
D_{n,t}^{\mathrm{base}}=[\mathcal D_{\mathcal W_n}^{\mathrm{base}}]_t,\qquad
D_{n,t}^{\mathrm{gen}}=[\mathcal D_{\mathcal W_n}^{\mathrm{gen}}]_t .
\label{eq:method_depth_triplet}
\end{equation}
The first is the metric camera-space \(z\)-depth in calibrated scene units.
The other two are normalized controls with no physical units. Only the first
may drive projection, visibility, correspondence, or trajectory evaluation;
only the last is supplied to the video generator.

Accepted tail evidence \(\mathcal E_n\) can complete only the newly exposed,
under-supported regions of the current generator control. The transferred
evidence is aligned to the fixed control and applied through
\begin{equation}
\mathcal D_{\mathcal W_n}^{\mathrm{gen}}
=(1-W_n)\odot\mathcal D_{\mathcal W_n}^{\mathrm{base}}
+W_n\odot\overline{\mathcal D}_{\mathcal W_n}(\mathcal E_n),
\label{eq:generative_depth_update}
\end{equation}
where \(W_n\) is zero outside the pointwise eligibility tests. These tests use
only \(\mathcal G_n\) and the evidence accepted before the current sampling
call; a missing overlap, a degenerate affine fit, or a failed agreement test
sets the affected weights to zero and preserves the base control exactly.
Exact alignment rules and thresholds are given in the supplementary appendix,
Sec.~A.3 (``Separated Depth Controls''). This branch changes visual conditioning
only: it never writes \(D^{\mathrm{geo}}\), collision geometry, or physical
state.

\begin{figure}[t]
    \centering
    \includegraphics[pagebox=mediabox,width=\linewidth,trim=12pt 53pt 50pt 83pt,clip]{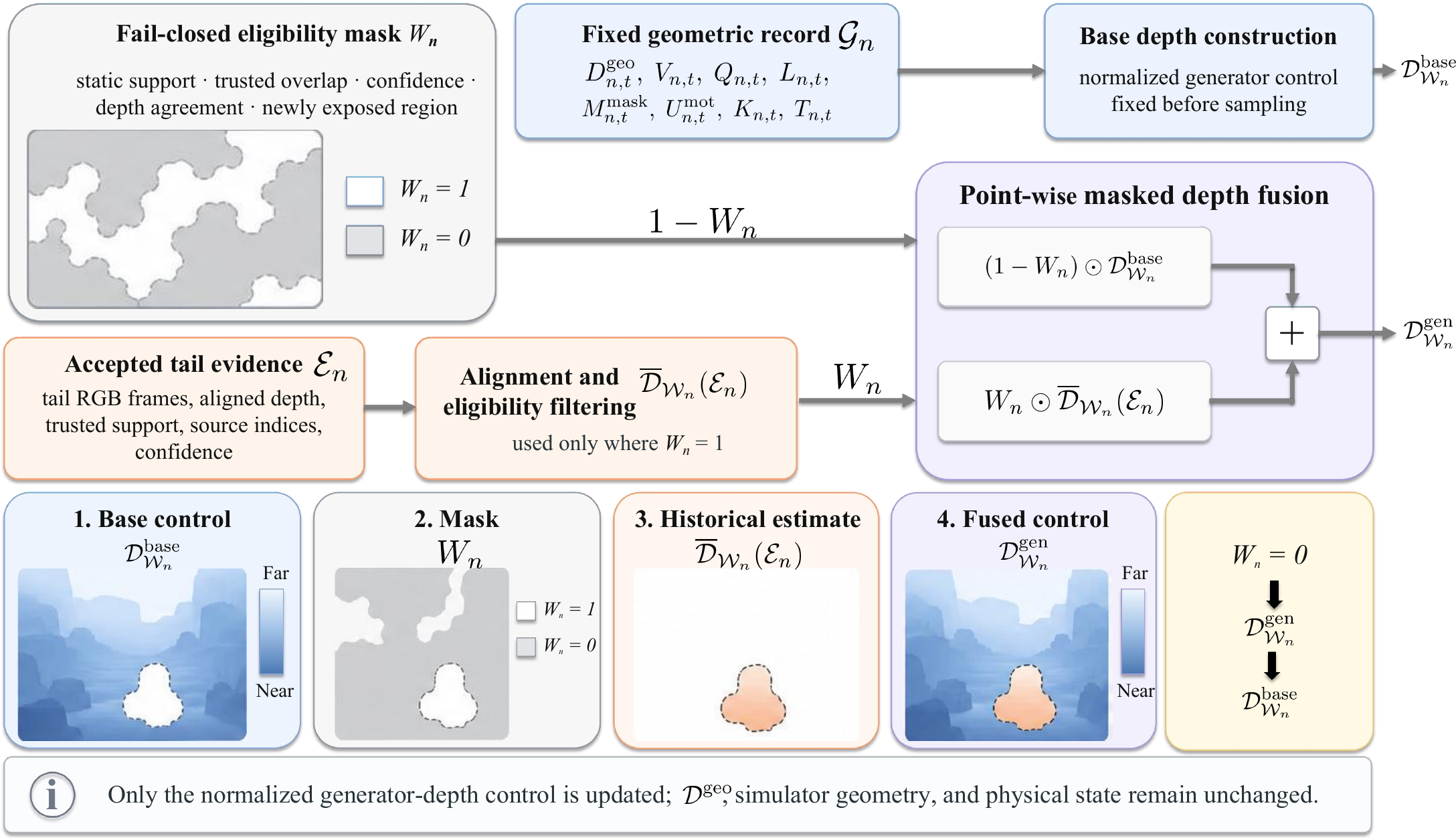}
    \caption{Generator-facing depth update, following the reference-style
    routing diagram. The fixed geometric record constructs the base control;
    accepted tail evidence is aligned and filtered into
    \(\overline{\mathcal{D}}(\mathcal E_n)\), while \(W_n\) gates the newly exposed,
    under-supported region. The final control is formed by point-wise fusion;
    outside the mask, the base control is preserved exactly. The depth maps
    are illustrative visualizations of the update, not additional simulator
    geometry.}
    \label{fig:depth_control_update}
\end{figure}

Figure~\ref{fig:depth_control_update} visualizes the four terms and the
point-wise routing: the fixed base control, aligned accepted tail evidence,
the eligibility mask, and the fused generator control.

The eligibility test is deliberately fail-closed. A candidate tail frame must
achieve at least \(0.20\) trusted static overlap, a Spearman agreement of at
least \(0.70\), a model-space NRMSE of at most \(0.08\), and a metric-space
NRMSE of at most \(0.12\). Completion is considered only when the median
newly visible fraction is at least \(0.05\), the target hole fraction is at
least \(0.01\), and the trusted source coverage is at least \(0.20\); at
least two of the configured tail frames must pass. Affine alignment uses the
overlap samples and is discarded when the fit is degenerate.

The mask \(W_n\) is constructed after these frame-level tests and is zero on
metric-supported or dynamic pixels. A temporal union followed by a small
dilation yields one stable support mask for the entire 81-frame window,
rather than a separately moving hole boundary in each frame. Accepted
evidence stores only aligned generator-depth estimates, support indices, and
confidence statistics. It is never interpreted as a new measurement of
simulator depth.

\subsection{Observation Synthesis and Staged Update}\label{sec:method_observation}

Once \(\mathcal G_n\) and \(\mathcal D_{\mathcal W_n}^{\mathrm{gen}}\) are
fixed, the finite-window diffusion model synthesizes the observation
\begin{equation}
Y_{\mathcal W_n}\sim p_\theta\!\left(
Y_{\mathcal W_n}\mid
\mathcal G_n,\mathcal D_{\mathcal W_n}^{\mathrm{gen}},I_0,\mathcal K_n
\right).
\label{eq:method_observation}
\end{equation}
The physical trajectory, camera poses, geometric record, and generator
controls remain fixed throughout this call. The permanent image anchor fixes
appearance, the current trajectory determines layout, and the committed
memory supplies only accepted historical appearance.

Controls are prepared at \(640\mathord{\times}352\), while the decoded output
is \(832\mathord{\times}480\) at \(16\) fps. An 81-frame window contains one
shared conditioning boundary and 80 new frames; consecutive windows therefore
advance by 80 decoded frames. The finite native cache handles adjacent
temporal continuity, whereas the geometry-routed pages described below
provide sparse long-range access, complementing the temporal prediction and
interpolation mechanisms used in prior video models~\cite{tpami_video_prediction_review,tpami_memcnet}.
The abstract state records the long-range
pages as \(\mathcal K_n\); the native cache is checkpointed together with the
same observation context. Model weights, the denoising schedule, and the
camera/physics controls remain frozen during this call.

The conditioning record is assembled before sampling and is not recomputed
from intermediate denoising states. In particular, simulator renderings serve
as control and evaluation signals rather than replacement RGB. This
distinction prevents a successful-looking simulator frame from silently
becoming the generated observation and keeps the observation branch from
acquiring physical authority.

The memory branch retains reliable static support from earlier committed
views. Geometry-routed correspondence uses \(D^{\mathrm{geo}}\), camera
parameters, visibility, confidence, and static-support masks to select at
most one eligible source page for each target location. Moving or deforming
proxies are excluded. The design is related to memory-based video
segmentation and tracking systems that use explicit temporal correspondence,
but here the eligibility test is additionally tied to persistent scene
geometry~\cite{tpami_vos_memory,tpami_tracking_survey}. Only visible,
high-confidence static support enters the long-range memory:
\begin{equation}
\Omega_t=V_t\cap M_t^{\mathrm{sta}}
\cap\{Q_t\geq\tau_q\}\cap\neg\Gamma_t^{\mathrm{edge}} ,
\label{eq:method_memory_support}
\end{equation}
where \(M_t^{\mathrm{sta}}\) is derived from the declaration and the
motion/deformation field, and \(\Gamma_t^{\mathrm{edge}}\) removes unreliable
depth boundaries. For a target pixel \(p\), the metric depth back-projects it
into the shared scene frame, and an earlier source view receives the
corresponding projection. Here \(T_t\) denotes the world-to-camera transform,
and \(\pi^{-1}\) returns a camera-frame point with the supplied metric
\(z\)-depth:
\begin{equation}
x_w=T_t^{-1}\pi^{-1}(p,D_t^{\mathrm{geo}}(p);K_t),
\qquad
(\widehat p_s,\widehat z_s)=\pi(T_sx_w;K_s).
\label{eq:method_cross_view_projection}
\end{equation}
The source must be in bounds, visible, static, and depth-consistent. Its
confidence is
\begin{equation}
c_s(p)=Q_t(p)Q_s(\widehat p_s)
\exp\!\left(-\frac{|D_s^{\mathrm{geo}}(\widehat p_s)-\widehat z_s|}
{\epsilon_s(p)}\right),
\label{eq:method_correspondence}
\end{equation}
where \(\epsilon_s(p)>0\) is the fixed metric-depth tolerance. The binary
mask \(m_s(p)\) equals one only when all tests pass. Among the earlier
committed pages, we select the one with the highest score, computed from
valid coverage and mean confidence; a target reads at most one page.

If \(A^0\) denotes the native reference-conditioned attention output and
\(A^{\mathrm{hist}}\) the output after reading an eligible page, the
historical read is introduced as the bounded residual
\begin{equation}
\begin{aligned}
A(p)&=A^0(p)+g_s(p)
       \left(A^{\mathrm{hist}}(p)-A^0(p)\right),\\
g_s(p)&=\lambda_{\mathrm{mem}}m_s(p)
       \operatorname{clip}(c_s(p),0,1),
\qquad 0\leq\lambda_{\mathrm{mem}}\leq1 .
\end{aligned}
\label{eq:gated_residual}
\end{equation}
Here \(m_s\) is the binary geometry-eligibility mask, and \(c_s\) combines
target/source confidence with metric-depth agreement. When no valid page
exists, \(g_s=0\) and the native reference-conditioned path is recovered
exactly.

The denoiser exposes clean attention features
\(Z_{\mathcal W_n}^{\mathrm{attn}}\) for selected source latents by
re-encoding them at the clean diffusion endpoint. The stored tensor is
therefore an attention-value page rather than a decoded RGB frame; the
operation changes only the attention features and does not copy pixels,
alter the target layout, or mix cache features with simulator visibility
fields. The staged page update is
\begin{equation}
\widehat{\mathcal K}_{n+1}
=\mathcal K_n\oplus
\Phi_{\mathrm{page}}\!\left(
Z_{\mathcal W_n}^{\mathrm{attn}},\mathcal G_n,\Omega_{\mathcal W_n}\right),
\label{eq:staged_page_update}
\end{equation}
where \(\oplus\) appends pages under their source-frame or source-latent IDs,
and \(\Omega_{\mathcal W_n}=\{\Omega_t:t\in\mathcal W_n\}\) is the
corresponding reliable static support. Tail evidence is staged from the
configured tail RGB frames and remains private, together with the pages,
until the quality gate accepts the whole window.

For the evaluated profile, the tail set is
\(\mathcal T_n=\{64,68,72,76,80\}\). Its candidate evidence is summarized as
\begin{equation}
\widehat{\mathcal E}_{n+1}
=\Phi_{\mathcal E}\!\left(
Y_{\mathcal W_n}|_{\mathcal T_n},\mathcal G_n,\mathcal R\right),
\label{eq:method_staged_evidence}
\end{equation}
where the operator stores only the aligned generator-depth estimates,
trusted support, source indices, and confidence statistics. The
simulator-consistent record is passed to this operator for testing purposes
only, never for overwriting.

The page and tail branches are both private candidate state. A page created
by the current window cannot be selected by another frame within that same
window, and tail evidence cannot affect the control tensor already used to
sample the current window. Their earliest possible effect occurs in the next
accepted window. This ordering renders the memory causal and the
depth-completion branch reproducible under retry.

\begin{figure}[!t]
    \centering
    \includegraphics[pagebox=mediabox,width=\linewidth,trim=4pt 13pt 54pt 6pt,clip]{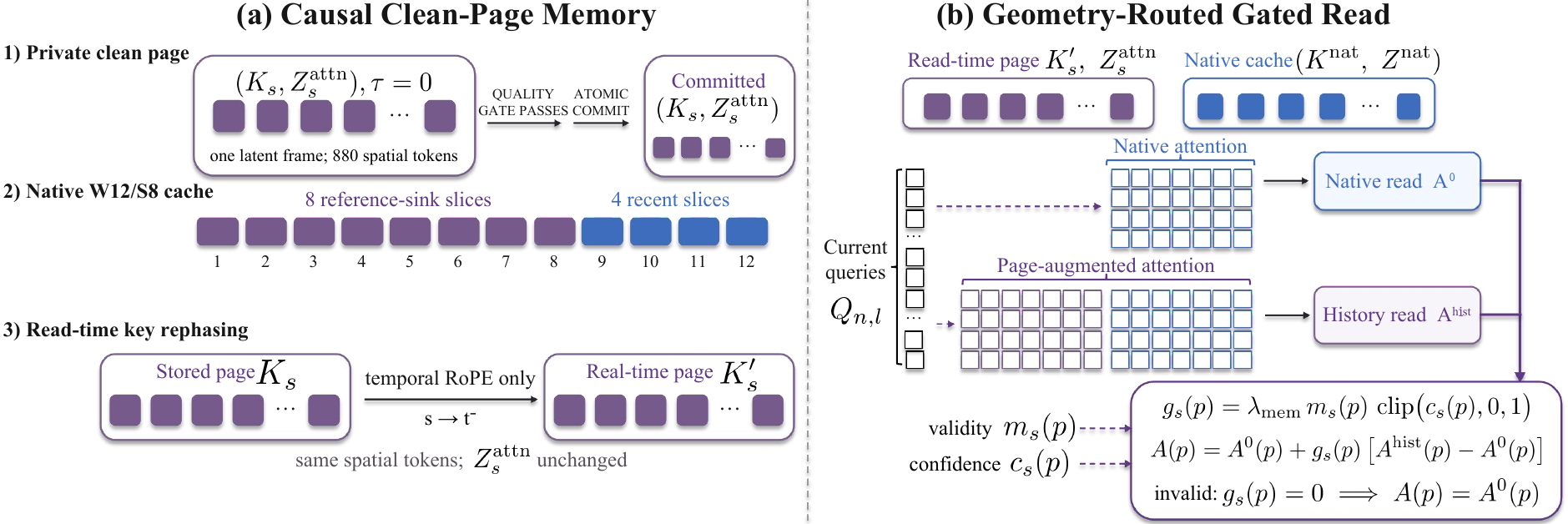}
    \caption{Reference-anchored appearance memory. Historical features become accessible only after their corresponding source window is committed. Simulator-consistent geometry determines whether a target location corresponds to reliable static support in an earlier view. In the diagram, \(\mathbf Z_s^{\mathrm{attn}}\) denotes the stored attention-value tensor and is distinct from the geometric visibility \(V_s\). The selected historical read is blended with the native reference-conditioned output through a bounded confidence gate. Invalid correspondence recovers the native path exactly.}
    \label{fig:appearance_memory}
\end{figure}

\subsection{Quality-Gated Atomic Commit}\label{sec:method_commit}

The quality gate is a fail-closed publication check applied to the decoded
window. It examines the media contract together with the fixed boundary,
black-frame, and frozen-motion diagnostics; the exact definitions and
thresholds are given in the supplementary appendix, Sec.~A.5 (``Quality Gate,
Atomicity, and Retry Policy''). It is not a
replacement for the simulator trajectory metrics.

Let \(Y_0,\ldots,Y_{80}\) denote the decoded 8-bit RGB frames and \(B_n\) the
last committed frame. We define \(d_t=\delta(Y_t,Y_{t-1})\) for \(t>0\) and
compute
\begin{equation}
\begin{aligned}
\delta(A,B)
&=\frac{1}{3HW}
  \sum_{p,c}\left|A(p,c)-B(p,c)\right|,\\
e_{\mathrm{ov}}
&=\delta(Y_0,B_n),\\
r_{\mathrm{bd}}
&=\frac{e_{\mathrm{ov}}}
{\max\!\left\{
  \operatorname{median}(d_1,\ldots,d_8),\,10^{-6}
\right\}}.
\end{aligned}
\label{eq:method_gate_boundary}
\end{equation}
with \(e_{\mathrm{ov}}=0\) for the first window. The boundary-motion ratio
uses the median of the first eight adjacent-frame changes (with a
\(10^{-6}\) floor). We also measure the fraction of nearly black pixels and
the fraction of adjacent changes below \(0.2\), denoted \(r_{\mathrm{black}}\)
and \(r_{\mathrm{freeze}}\), respectively.

The reported decision is the conjunction
\begin{equation}
\begin{aligned}
q_n ={}&
\mathbf{1}[\mathrm{media}]
\mathbf{1}[e_{\mathrm{ov}}\leq 4]
\mathbf{1}[r_{\mathrm{bd}}\leq 3] \\
&\times
\mathbf{1}[r_{\mathrm{black}}\leq 0.10]
\mathbf{1}[r_{\mathrm{freeze}}\leq 0.25]
\chi_{\mathrm{repr}} .
\end{aligned}
\label{eq:method_quality_gate}
\end{equation}

where the media contract requires 81 decodable \(832\mathord{\times}480\)
frames. Here \(\chi_{\mathrm{repr}}=1\) unless a pre-commit
feedback-reprojection manifest is supplied, in which case it is the
indicator of \(e_{\mathrm{repr}}\leq24\). The manifest was absent in the
reported runs. Simulator RGB MAE is logged as a diagnostic and is not an
acceptance criterion.

For \(q_n=1\), the terminal physical and camera states, the staged
appearance pages, and the staged tail evidence are published together
according to Eq.~\eqref{eq:knowledge_transition}. For \(q_n=0\), the working
simulator/camera copy, the candidate pages, and the candidate evidence are
discarded, and \(\mathcal S_n\) remains unchanged; the candidate native cache
and the random-state checkpoint are restored as well. A retry preserves the
same committed index, action, trajectory, generator controls, and committed
memory; only the sampling seed may change. The next user action is released
only after \(q_n=1\), so a retry cannot advance the physical or epistemic
state. Consequently, generated RGB can extend appearance knowledge only
through an accepted update and never writes back to \(X_n\), \(C_n\), or
\(D^{\mathrm{geo}}\).

The gate runs after decoding and before any candidate cache state, page,
tail evidence, or terminal simulator state becomes persistent. Thresholds
are fixed for the attempt and are never relaxed; an explicit retry reuses
the committed checkpoint and all candidate controls, with only the sampling
seed permitted to differ. Failed attempts therefore cannot advance either
the state index or the history.

\begin{figure}[!t]
    \centering
    \includegraphics[pagebox=mediabox,width=\linewidth,trim=45pt 618pt 49pt 15pt,clip]{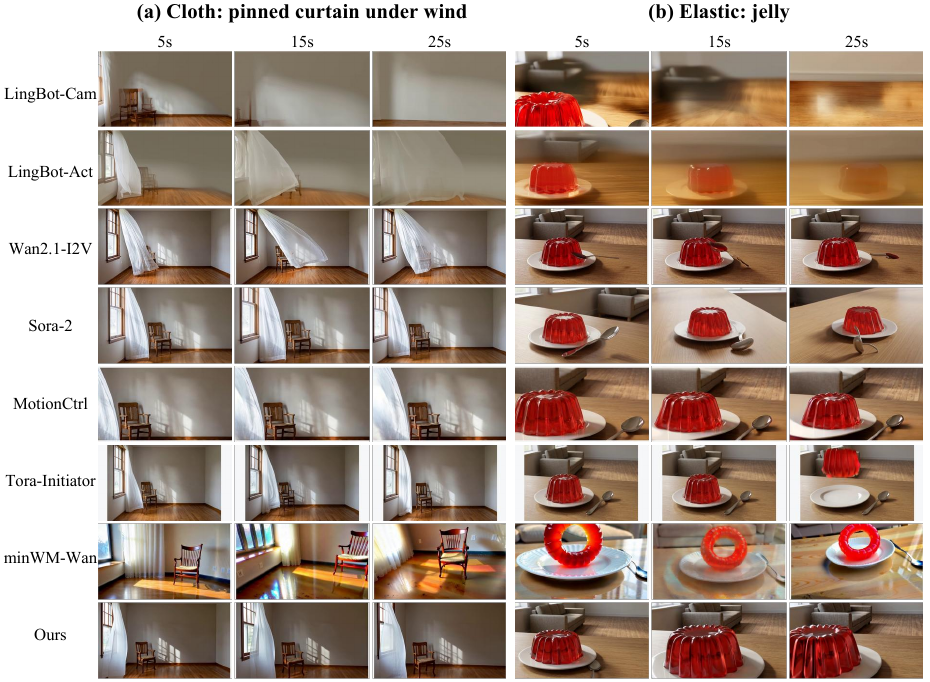}
    \caption{Full-duration key-frame comparison across all eight evaluated methods. Within each scene, methods are rows. Each complete baseline output is uniformly mapped to the corresponding 30.06\,s TourPhysics duration, and columns sample \(t=5,15,25\,\mathrm{s}\) on that matched timeline; no looping, tail padding, or per-output frame selection is used. (a) A pinned curtain under wind. (b) A jelly. Native durations and generation inputs follow Table~\ref{tab:input_capabilities}. Frames are uncropped and padded to a common aspect ratio, with no color correction or replacement.}
    \label{fig:quality_comparison}
\end{figure}

\section{Experiments}\label{sec:experiments}

\subsection{Experimental Setup}

The experiments are organized around the four capabilities required by the
problem formulation. The first experiment asks whether the generated camera
and object motion follows the fixed physical and camera trajectory. The second checks
whether contacts and deformations occur at the declared times and with the
declared shape changes. The third examines whether long-horizon generation
preserves the input scene and the intended camera motion. The final experiment
isolates the persistent cache and the geometry-routed appearance memory. This
order follows Figure~\ref{fig:framework}: the orange physical-transition branch
defines the target trajectory, the purple observation branch renders it, and the
green lower loop carries visual evidence across revisits.

\paragraph{Benchmark and execution protocol.}
The benchmark contains 108 simulator-defined interaction sequences initialized
from 61 source-image identities. It covers camera-only tours and manipulations
of rigid, cloth, elastic, granular, and fluid proxies. Each sequence starts
from the same single image and declarative configuration used to initialize the
scene hypothesis. For every action, the simulator exports a camera and object
trajectory together with the geometric record and a paired simulator rendering;
the video model generates an observation for that fixed window. The next action
is released only after the observation passes the window-level commit rule.
Thus, every paired comparison has a declared target trajectory and an explicit
commit boundary, rather than an unconstrained video continuation. The benchmark
measures agreement with this controllable hypothesis, which is the quantity
needed for exploration and manipulation from one image.

\paragraph{Compared systems and common measurements.}
The comparison spans the main conditioning regimes used for controllable video
and interactive world models. LingBot-Cam and LingBot-Act~\cite{robbyant2026lingbot}
represent camera- and action-conditioned open-source interactive world-model
variants; minWM-Wan~\cite{zhao2026minwm} represents an interactive video world
model. Wan2.1-I2V~\cite{wan2025open} and Sora-2~\cite{openai_sora2_2025} are
general image-to-video generators whose primary task is appearance and motion
synthesis, while MotionCtrl~\cite{wang2024motionctrl} and
Tora-Initiator~\cite{zhang2025tora} are controllable video generators driven by
poses or point trajectories. TourPhysics is the simulator-guided persistent model that receives
the declarative scene and computes the physical trajectory before synthesis.
Table~\ref{tab:input_capabilities} records the camera, depth, tracking, mask,
and physics inputs available to each system, including native and diagnostic
durations. This makes the comparison useful for the corresponding task: each
method is evaluated through its supported interface, while all outputs are
scored against the same simulator-defined targets. Paired object and physical
metrics use 8-fps videos at \(682\times384\) and the same frozen-mask
CoTracker3~\cite{karaev2024cotracker3,tpami_tracking_survey,tpami_tracking_experimental}
pipeline; camera and appearance protocols
are specified with their respective experiments below.

Table~\ref{tab:input_capabilities} summarizes the method-specific generation
inputs and native or diagnostic temporal support. It is an input-protocol
matrix, rather than a performance table: its purpose is to make the information
available to each comparison method explicit before interpreting the results.

\begin{table}[!t]
    \caption{
        Evaluation inputs and temporal support.
        ``Int.'' denotes controls derived internally from the declared simulator state,
        ``Text'' under Camera denotes prompt-only specification,
        and \(^{\mathrm{d}}\) marks a uniformly time-stretched full-timeline diagnostic.
    }
    \label{tab:input_capabilities}

    \centering
    \small
    \setlength{\tabcolsep}{3.5pt}
    \renewcommand{\arraystretch}{1.08}

    \begin{tabular}{@{}
        l
        C{0.050\linewidth}
        C{0.065\linewidth}
        C{0.125\linewidth}
        C{0.055\linewidth}
        C{0.085\linewidth}
        C{0.055\linewidth}
        C{0.105\linewidth}
        C{0.175\linewidth}
    @{}}
        \toprule
        Method
        & Image
        & Text
        & Camera
        & Depth
        & Track
        & Mask
        & \makecell{Physics\\script}
        & \makecell{Native / diagnostic\\duration} \\
        \midrule

        TourPhysics
        & \checkmark
        & Scene
        & \mbox{Action/pose}
        & Int.
        & Int.
        & Int.
        & Declarative
        & \mbox{25.1/30.1\,s} \\

        LingBot-Cam
        & \checkmark
        & Frozen
        & Pose
        & --
        & --
        & --
        & --
        & \mbox{25.1/30.1\,s} \\

        LingBot-Act
        & \checkmark
        & Frozen
        & \mbox{Discrete action}
        & --
        & --
        & --
        & --
        & \mbox{25.1/30.1\,s} \\

        Wan2.1-I2V
        & \checkmark
        & Frozen
        & Text
        & --
        & --
        & --
        & --
        & \mbox{10.7\,s / 25--30\,s\(^{\mathrm{d}}\)} \\

        Sora-2
        & \checkmark
        & Frozen
        & Text
        & --
        & --
        & --
        & --
        & \mbox{12.2\,s} \\

        MotionCtrl
        & \checkmark
        & --
        & \mbox{RT pose}
        & --
        & --
        & --
        & --
        & \mbox{1.4\,s / 25--30\,s\(^{\mathrm{d}}\)} \\

        Tora-Initiator
        & \checkmark
        & Frozen
        & --
        & --
        & \mbox{Oracle 2D}
        & --
        & --
        & \mbox{6.1\,s} \\

        minWM-Wan
        & \checkmark
        & Frozen
        & \mbox{Discrete action}
        & --
        & --
        & --
        & --
        & \mbox{39.8\,s} \\

        \bottomrule
    \end{tabular}

    \vspace{3pt}

    \begin{minipage}{0.98\linewidth}
        \footnotesize
        Tora receives the simulator-projected initiator trajectory.
        Depth, track, mask, and physics-script columns are generation-time inputs.
        The \(^{\mathrm{d}}\) rows use timestamp scaling without looping or tail padding.
    \end{minipage}
\end{table}

\subsection{Camera and Object Motion}

This experiment tests the most direct consequence of fixing the trajectory
before observation synthesis: the camera should execute the prescribed tour and
the visible objects should follow the simulator state. It evaluates the camera
path and object-motion outputs of the orange ``Physical transition'' branch in
Figure~\ref{fig:framework}, rather than judging motion only by visual
smoothness.

\paragraph{Protocol and metrics.}
Camera scores use the 49 sequences from 40 identities whose reference poses are
exact runtime exports. ViPE camera centers are timestamp-matched to the
simulator centers, and one Umeyama similarity transform is fitted over each
complete sequence. ATE is the RMS camera-center deviation after this alignment,
so lower values mean closer spatial tracking. PLR is the recovered-to-declared
path-length ratio, with one indicating equal traveled distance. T-Dir and R-Dir
are the percentages of translation-only and rotation-only schedule segments
whose recovered motion points to the correct half-space. These four metrics
separate absolute position error, path scaling, and action-direction
correctness.

Object scores use paired simulator and generated videos sampled at 8 fps and
\(682\times384\). Frozen initial masks seed the same CoTracker3 multi-point
tracker, and an object is retained only when its centers are jointly visible for
at least eight frames and 15\% of the sequence. ADE is the mean normalized
center error, End Err. is the error at the last jointly visible frame, S@0.05 is
the fraction of objects whose ADE is within 5\% of the image diagonal, and Vis.
is the jointly visible fraction. TourPhysics contributes 168 tracked objects in
93 sequences from 49 identities; the baselines use the valid records listed in
the table footnote. Complete alignment and support definitions are provided in
the supplementary appendix, Sec.~B.2 (``Camera and Object Motion'').

\begin{table}[!t]
    \caption{
        Prescribed camera- and object-motion adherence.
        Camera metrics compare generated and declared trajectories;
        object metrics measure image-space agreement with paired simulator-rendered sequences.
        Generation inputs are listed in Table~\ref{tab:input_capabilities}.
    }
    \label{tab:camera_object_motion_comparison}

    \centering
    \small
    \setlength{\tabcolsep}{3.2pt}
    \renewcommand{\arraystretch}{1.08}

    \begin{tabular}{@{}
        L{0.135\linewidth}  
        C{0.070\linewidth}  
        C{0.082\linewidth}  
        C{0.078\linewidth}  
        C{0.088\linewidth}  
        C{0.068\linewidth}  
        C{0.105\linewidth}  
        C{0.082\linewidth}  
        C{0.070\linewidth}  
    @{}}
        \toprule

        \multirow{2}{*}{Method}
        & \multicolumn{4}{c}{Camera}
        & \multicolumn{4}{c}{Object Motion} \\

        \cmidrule(lr){2-5}
        \cmidrule(lr){6-9}

        & ATE $\downarrow$
        & PLR $\rightarrow 1$
        & T-Dir $\uparrow$
        & R-Dir $\uparrow$
        & ADE $\downarrow$
        & End Err. $\downarrow$
        & S@0.05 $\uparrow$
        & Vis. $\uparrow$ \\

        \midrule

        LingBot-Cam
        & 0.1031
        & 4.8434
        & 88.54\%
        & 53.00\%
        & 0.0833
        & 0.1605
        & 18.37\%
        & 35.04\% \\

        LingBot-Act
        & 0.1235
        & 6.3812
        & 68.75\%
        & 61.00\%
        & 0.0834
        & 0.1540
        & 27.09\%
        & 50.56\% \\

        Wan2.1-I2V
        & 0.1436
        & 3.8702
        & 60.42\%
        & 42.00\%
        & 0.0774
        & 0.1233
        & 31.61\%
        & 76.43\% \\

        Sora-2
        & 0.0560
        & 2.3898
        & 78.52\%
        & 99.99\%
        & 0.0721
        & 0.1419
        & 43.47\%
        & 81.88\% \\

        MotionCtrl
        & 0.0783
        & \textbf{0.8517}
        & 50.35\%
        & 38.00\%
        & 0.0871
        & 0.1474
        & 19.74\%
        & \textbf{88.46\%} \\

        Tora-Initiator
        & N/A
        & N/A
        & N/A
        & N/A
        & 0.0462
        & 0.0865
        & 70.05\%
        & 80.10\% \\

        minWM-Wan
        & 0.1092
        & 1.9820
        & 89.58\%
        & 46.00\%
        & 0.1051
        & 0.2213
        & 9.46\%
        & 69.63\% \\

        \midrule

        Ours
        & \textbf{0.0233}
        & 1.4301
        & \textbf{98.26\%}
        & \textbf{93.00\%}
        & \textbf{0.0297}$^{\ddagger}$
        & \textbf{0.0524}$^{\ddagger}$
        & \textbf{81.18\%}$^{\ddagger}$
        & 75.20\%$^{\ddagger}$ \\

        \bottomrule
    \end{tabular}

    \vspace{3pt}

    \begin{minipage}{0.98\linewidth}
        \footnotesize
        $^{\ddagger}$TourPhysics object metrics cover 168 objects in 93 sequences
        from 49 input identities after excluding trajectories with less than
        15\% joint visibility.
        Camera metrics use 49 sequences from 40 identities with runtime-exact poses.
        Sora R-Dir has one eligible segment and is excluded from best-result highlighting.
        Wan and MotionCtrl are uniformly time-stretched to full duration.
        Bold marks the best displayed result on each method's valid support.
    \end{minipage}

\end{table}

Table~\ref{tab:camera_object_motion_comparison} shows that TourPhysics has the
lowest camera ATE (0.0233) and the highest translation- and rotation-direction
hits (98.26\% and 93.00\%). It also has the lowest object ADE (0.0297) and
endpoint error (0.0524), with 81.18\% of retained objects within the 5\%
diagonal threshold. These are the measurements most closely tied to the
declared action and state transition: the simulator fixes the path first, and
the generated observation preserves that path when it is rendered. PLR and
Vis. provide complementary information about total path scaling and track
support, rather than replacing the spatial and directional adherence measures.

\subsection{Adherence to Simulator-Defined Interactions}

The second experiment asks whether a visually plausible interaction also
preserves the simulator-defined event sequence. It uses the same paired
simulator/generated videos and the orange physical-transition branch in
Figure~\ref{fig:framework}, but evaluates event timing, collision response, and
deformation rather than whole-object centers.

\paragraph{Protocol and metrics.}
The simulator supplies the contact onset, initiator-target pair, and declared
deformation interval for each sequence. The generated and simulator videos are
tracked with the frozen-mask CoTracker3 protocol at 8 fps; records are reduced
within a sequence and then averaged equally over sequences, so a sequence with
several objects or events does not dominate. Contact Err. is the timing error
between the generated and simulator contact proxies. Response Acc. is the
percentage of observable targets that exhibit the prescribed post-contact
motion. Overlap Exc. is the excess 90th-percentile 2D hull overlap after
contact. Shape Err. is the normalized Procrustes point error for a deformable
object, Area Drift is the deviation of its projected area from the initial
ratio, and Recovery Acc. is the percentage of declared deformation events that
meet both the amplitude and post-event recovery criteria. The auxiliary SA and
PC scores from VideoPhy-2~\cite{bansal2025videophy2} measure semantic adherence
and physical commonsense on a 1--5 scale; their average is a perceptual
diagnostic, not a trajectory score. The event definitions and effective support
for each method are given in the supplementary appendix, Sec.~B.3
(``Simulation-Trajectory Adherence'').
```latex

\begin{table}[!t]
    \caption{
        Adherence to simulator-defined physical trajectories.
        Collision and deformation columns are tracked image-space proxies
        computed against paired simulator-rendered videos.
        VideoPhy-2 supplies an auxiliary perceptual diagnostic.
    }
    \label{tab:physics_comparison}

    \centering
    \small
    \setlength{\tabcolsep}{2.8pt}
    \renewcommand{\arraystretch}{1.08}

    \begin{tabular}{@{}
        L{0.130\linewidth}
        C{0.092\linewidth}
        C{0.090\linewidth}
        C{0.086\linewidth}
        C{0.080\linewidth}
        C{0.078\linewidth}
        C{0.092\linewidth}
        C{0.052\linewidth}
        C{0.052\linewidth}
        C{0.052\linewidth}
    @{}}
        \toprule

        \multirow{2}{*}{Method}
        & \multicolumn{3}{c}{Collision Proxies}
        & \multicolumn{3}{c}{Deformation Proxies}
        & \multicolumn{3}{c}{\makecell{VideoPhy-2 AutoEval\\(Auxiliary)}} \\

        \cmidrule(lr){2-4}
        \cmidrule(lr){5-7}
        \cmidrule(lr){8-10}

        & \makecell{Contact\\Err. $\downarrow$}
        & \makecell{Response\\Acc. $\uparrow$}
        & \makecell{Overlap\\Exc. $\downarrow$}
        & \makecell{Shape\\Err. $\downarrow$}
        & \makecell{Area\\Drift $\downarrow$}
        & \makecell{Recovery\\Acc. $\uparrow$}
        & SA $\uparrow$
        & PC $\uparrow$
        & Avg $\uparrow$ \\

        \midrule

        LingBot-Cam
        & \mbox{6.9196\,s}
        & $<0.01\%$
        & \textbf{$<10^{-4}$}
        & \textbf{0.0664}
        & 0.5136
        & 50.00\%
        & 2.9444
        & 4.4167
        & 3.6806 \\

        LingBot-Act
        & \mbox{7.8304\,s}
        & $<0.01\%$
        & \textbf{$<10^{-4}$}
        & 0.0809
        & 0.1757
        & $<0.01\%$
        & 2.9352
        & 4.2315
        & 3.5833 \\

        Wan2.1-I2V
        & \mbox{5.1875\,s}
        & \textbf{33.33\%}
        & \textbf{$<10^{-4}$}
        & 0.0866
        & 0.2763
        & 30.00\%
        & 2.9444
        & 4.0185
        & 3.4815 \\

        Sora-2
        & \mbox{4.9107\,s}
        & \textbf{33.33\%}
        & \textbf{$<10^{-4}$}
        & 0.1351
        & 0.3369
        & 33.33\%
        & \textbf{2.9630}
        & 4.2778
        & 3.6204 \\

        MotionCtrl
        & \mbox{7.0893\,s}
        & $<0.01\%$
        & \textbf{$<10^{-4}$}
        & 0.1234
        & 0.3539
        & N/A
        & 2.9352
        & \textbf{4.7963}
        & \textbf{3.8657} \\

        Tora-Initiator
        & \mbox{5.8839\,s}
        & $<0.01\%$
        & \textbf{$<10^{-4}$}
        & 0.1081
        & 0.2249
        & $>99.99\%$
        & 2.8660
        & 4.2784
        & 3.5722 \\

        minWM-Wan
        & \mbox{4.5639\,s}
        & 22.92\%
        & \textbf{$<10^{-4}$}
        & 0.1530
        & 0.5046
        & $<0.01\%$
        & 2.9352
        & 4.1296
        & 3.5324 \\

        \midrule

        Ours
        & \textbf{3.3304\,s}$^{\mathrm{C}}$
        & 16.67\%$^{\mathrm{C}}$
        & \textbf{$<10^{-4}$}$^{\mathrm{C}}$
        & 0.0704$^{\mathrm{M}}$
        & \textbf{0.0866}$^{\mathrm{M}}$
        & \textbf{64.29\%}$^{\mathrm{R}}$
        & 2.9167
        & 4.3148
        & 3.6157 \\

        \bottomrule
    \end{tabular}

    \vspace{3pt}

    \begin{minipage}{0.98\linewidth}
        \footnotesize
        ``N/A'' denotes an incompatible protocol.
        $^{\mathrm{C}}$Contact, response, and overlap cover 14/5/9 valid
        TourPhysics sequences from 6/3/5 input identities;
        $^{\mathrm{M}}$shape and area cover 30 records in 25 sequences
        from 13 identities;
        $^{\mathrm{R}}$recovery covers eight events in seven sequences
        from two identities.
        Tora recovery has one eligible event and is excluded from
        best-result highlighting.
        All deterministic columns use the same frozen CoTracker3 pipeline,
        with method-specific support after visibility filtering;
        bold marks the best displayed value on that method-specific support
        rather than on a common subset.
        VideoPhy-2 covers 108 sequences (Tora: 97) and reports custom-data
        mean SA/PC on a 1--5 scale; Avg is their arithmetic mean.
        Wan and MotionCtrl are time-stretched to full duration.
        Zero displayed 2D overlap does not imply zero 3D penetration.
    \end{minipage}

\end{table}

Table~\ref{tab:physics_comparison} shows the strengths of the simulator-guided
branch on the deterministic interaction measures. TourPhysics reaches the
lowest contact-timing error (3.3304 s), the lowest area drift (0.0866), and the
highest recovery accuracy (64.29\%); its shape error (0.0704) is second only to
LingBot-Cam on that supported subset. The contact and deformation columns answer
different parts of the same question: the first group tests when an event
happens, while the second tests whether the generated object changes in the
right way and returns toward its post-event state. The perceptual VideoPhy-2
columns are reported alongside these deterministic measures as an auxiliary
appearance diagnostic; the deterministic columns provide the trajectory result.

\subsection{Appearance and Camera-Control Diagnostics}

\begin{figure}[!t]
    \centering
    \includegraphics[pagebox=mediabox,width=\linewidth]{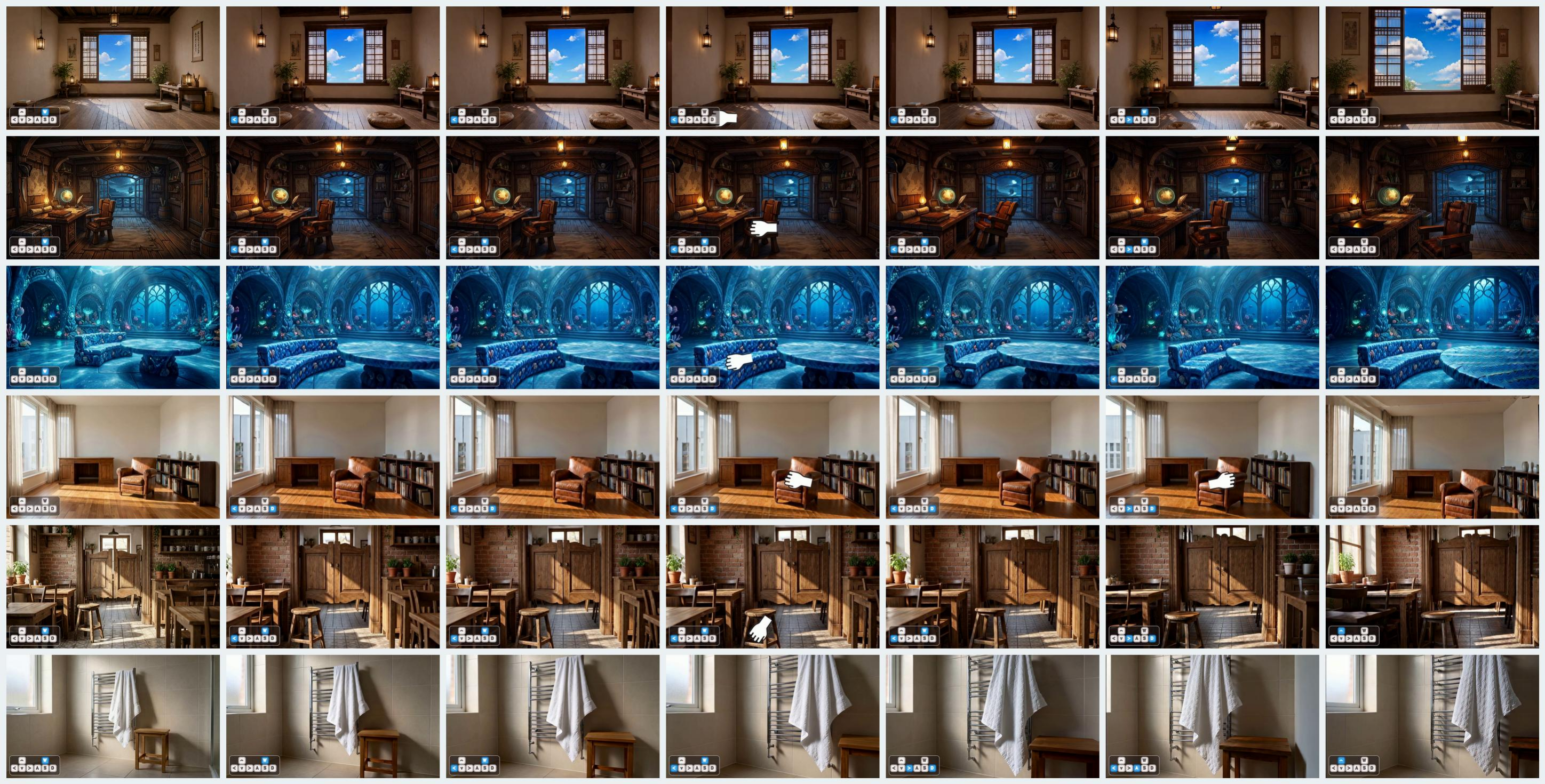}
    \caption{TourPhysics results across several scenarios. Key frames progress from left to right and show long-horizon appearance, camera motion, and physical interactions.}
    \label{fig:quality_results}
\end{figure}

The trajectory experiments establish control, while this experiment checks whether
that control is accompanied by stable visual content. Figure~\ref{fig:quality_comparison}
uses all eight systems on two representative scenes, a pinned curtain under
wind and a jelly. Each complete baseline output is mapped to the 30.06-s
TourPhysics timeline and sampled at 5, 15, and 25 s, with no looping, tail
padding, or per-output frame selection. Figure~\ref{fig:quality_results} extends
the view to several long-horizon camera and interaction scenarios.

The quantitative diagnostics measure complementary properties of the generated
observations. Input-video subject consistency measures whether the foreground
appearance remains aligned with the source video, background consistency tests
the static scene, and camera-motion classification accuracy measures whether the
generated motion retains the intended camera-motion class. TourPhysics obtains
0.9743, 0.9791, and 0.8921 on these three quantities, respectively. These
diagnostics connect the purple observation branch and the green memory loop in
Figure~\ref{fig:framework} to the visible long-horizon behavior: the model can
follow a physical/camera trajectory while retaining the reference scene. The
underlying masks, aggregation, and temporal normalization are specified in the
supplementary appendix, Sec.~B.4 (``Appearance and Long-Horizon Consistency'').

\begin{table}[!t]
    \caption{
        Long-horizon observation and appearance-memory diagnostics.
        Cache drift is measured on paired long-video sequences.
        Memory changes are reported relative to the same generator with
        appearance memory disabled; negative values denote lower error.
    }
    \label{tab:memory_diagnostics}

    \centering
    \small
    \setlength{\tabcolsep}{4.0pt}
    \renewcommand{\arraystretch}{1.08}

    \begin{tabular}{@{}
        L{0.300\linewidth}
        C{0.120\linewidth}
        C{0.120\linewidth}
        C{0.120\linewidth}
        C{0.120\linewidth}
        C{0.120\linewidth}
    @{}}
        \toprule

        Variant
        & Scope
        & \makecell{End-to-start\\drift $\downarrow$}
        & \makecell{Tail-to-head\\drift $\downarrow$}
        & \makecell{History error\\change $\downarrow$}
        & \makecell{Known-region\\change $\downarrow$} \\

        \midrule

        Ordinary temporal cache
        & \mbox{72 sequences}
        & 26.33
        & 20.38
        & --
        & -- \\

        Persistent sink cache
        & \mbox{72 sequences}
        & \textbf{11.17}
        & \textbf{7.74}
        & --
        & -- \\

        Appearance memory, development
        & \mbox{4 identities}
        & --
        & --
        & \textbf{$-8.25\%$}
        & \textbf{$-2.99\%$} \\

        Appearance memory, held-out
        & \mbox{14 identities}
        & --
        & --
        & \textbf{$-2.39\%$}
        & \textbf{$-0.90\%$} \\

        \bottomrule
    \end{tabular}

    \vspace{3pt}

\end{table}

\subsection{Long-Horizon Cache and Appearance Memory}

The final experiment targets the failure mode that motivates persistent memory:
a camera can revisit a surface after many intervening actions, and a finite
temporal cache alone does not preserve its appearance. It isolates the green
lower loop and long-horizon filmstrip in Figure~\ref{fig:framework} in two
stages. The cache comparison uses 72 paired long videos from 27 identities and
keeps the observation backbone and generator fixed while replacing an ordinary
finite cache with the persistent sink cache. The memory comparison uses four
development identities and an identity-disjoint held-out set of 14 dynamic
identities. Each selected sequence has six windows, 481 output frames, 51
planned target latents, and revisit chunks 3--5; physical and camera trajectories,
generator-facing depth, checkpoints, prompts, and random seeds are matched
between memory-on and memory-off runs.

The cache metrics are end-to-start drift and tail-to-head drift, both computed
from the mean decoded 8-bit channel value, so lower values mean less long-range
brightness change. The memory metrics are relative changes in historical-region
error on routed, depth-consistent correspondences and known-region error on the
simulator-consistent background; negative percentages indicate lower error with
memory. This separation follows the distinction between temporal association
and appearance filling emphasized by memory-based video segmentation and
temporally-aware inpainting~\cite{tpami_vos_memory,tpami_video_inpainting}.
Formula definitions, masks, aggregation, and the identity-disjoint selection
protocol are provided in the supplementary appendix, Sec.~B.4 (``Appearance
and Long-Horizon Consistency'').

Table~\ref{tab:memory_diagnostics} separates the two mechanisms cleanly. The
persistent sink cache reduces end-to-start drift from 26.33 to 11.17 and
tail-to-head drift from 20.38 to 7.74. Geometry-routed appearance memory then
reduces historical-region error by 8.25\% on development and 2.39\% on the
held-out identities, while known-region error changes by $-2.99\%$ and
$-0.90\%$. The first comparison shows why the cache is needed for long windows;
the second shows that accepted, geometry-routed history improves revisits while
leaving the physical and camera trajectory and generator-facing depth unchanged.

\section{Conclusion}\label{sec:concl}

We presented TourPhysics, an online framework for exploring and manipulating a controllable scene hypothesis initialized from a single image and a declarative physical configuration. Its architecture assigns separate roles to simulator state, simulator-consistent geometry, generator controls, and accepted appearance memory.

For each action, TourPhysics computes the physical and camera trajectory before generating an observation. Accepted observations publish the terminal state and update later appearance memory and generator-facing depth, while the committed state and geometry remain fixed during synthesis and retry. A permanent image reference and geometry-routed memory preserve appearance when the camera revisits earlier surfaces.

On simulator-defined camera tours and manipulations, TourPhysics improves alignment with prescribed camera and object trajectories, preserves the input scene, and reduces long-horizon appearance drift. Future work will study real-world physical evaluation, more complete scene geometry, and fixed-capacity memory for longer tours.


\clearpage
\appendix
\makeatletter
\@addtoreset{equation}{section}
\makeatother
\renewcommand{\theequation}{\thesection\arabic{equation}}
\section{Implementation and Evaluation Details}
\label{app:details}

This appendix summarizes implementation and evaluation details omitted from the
main paper. The single-image decomposition, explicit scene alignment, and
multi-physics initialization build on the corresponding components of the ACM
Multimedia 2026 conference paper PhysOmni~\cite{zhang2026physomni}. TourPhysics
retains this physical front end but extends it with persistent online state,
action-boundary trajectory simulation, transactional observation publication,
separated geometric and generative controls, and long-horizon appearance memory.

\subsection{Single-Image Scene Construction}
\label{app:scene_construction}

\paragraph{Instance decomposition and explicit proxies.}
Given the input image \(I_0\) and user-provided object descriptions, SAM
3~\cite{carion2025sam3segmentconcepts} extracts instance masks
\(\{M_i\}_{i=1}^{N}\). Each image--mask pair is then lifted by SAM 3D
Objects~\cite{chen2025sam} into an explicit triangular mesh
\(\mathcal S_i^{\mathrm{loc}}\) and an object-to-camera similarity transform
\((s_i,R_i,t_i)\). A local vertex is placed in the common camera frame by
\begin{equation}
    v_{i}^{\mathcal C}
    =s_iR_iv_{i}^{\mathrm{loc}}+t_i.
    \label{eq:app_object_lifting}
\end{equation}
These explicit surfaces provide the simulator's contact and collision
interfaces.
The foreground union is removed cumulatively with LaMa
inpainting~\cite{suvorov2021resolution,tpami_partialconv,tpami_video_inpainting}
to form a clean static background plate,
\begin{equation}
    B_0=\mathcal F_{\mathrm{inp}}
    \left(I_0,\bigcup_{i=1}^{N}M_i\right).
    \label{eq:app_background_inpaint}
\end{equation}
The unmodified \(I_0\) is retained separately as the permanent appearance
reference. The inpainted plate is used only for camera-induced background
exposure.

\paragraph{Gravity-aligned scene coordinates.}
Independently reconstructed meshes share an image-conditioned camera frame but
need a stable gravity direction and ground height before simulation. Let
\(\widehat u_{\mathcal C}\) be the provisional up direction. For each mesh, we
select the lowest \(\eta\)-quantile of vertices under
\(h(v)=v^{\mathsf T}\widehat u_{\mathcal C}\), and pool these likely contact
anchors into \(\mathcal P_{\mathrm{base}}\). We use \(\eta=0.05\) by default
and relax it to \(0.10\) when too few anchors are available. A robust plane is
then fitted as
\begin{equation}
    (n^*,d^*)=\arg\min_{\|n\|_2=1,d}
    \sum_{p\in\mathcal P_{\mathrm{base}}}
    \rho\!\left(n^{\mathsf T}p+d\right),
    \label{eq:app_ground_plane}
\end{equation}
where \(\rho\) is implemented by RANSAC inlier selection. Restricting the fit to
low object anchors prevents dense vertical structures from dominating the
estimated ground. A rigid transform rotates \(n^*\) onto the world \(z\)-axis,
recenters the pooled scene, and translates the lowest transformed anchor to
\(z=0\). Small residual pairwise overlaps are corrected by bounded translations
along the minimum-overlap bounding-box axis. These corrections are part of the
declared initial hypothesis, not online simulator updates.

\paragraph{Reference-camera and depth alignment.}
Monocular depth and camera estimates provide the initial reference-view
geometry. We refine the camera by rendering the union silhouette of the aligned
meshes and maximizing its overlap with \(\bigcup_iM_i\). A coarse-to-fine
coordinate search translates the camera and look-at point together along the
forward, right, and up axes, followed by a bounded image-plane correlation
refinement. For visible pixels with valid monocular and simulator depth, we fit
an affine alignment using the same scale and shift \((\alpha,\beta)\) as in the
main text:
\begin{equation}
    D^{\mathrm{geo}}_0(p)
    =
    \alpha\widetilde D_0(p)+\beta,
    \label{eq:app_depth_alignment}
\end{equation}
where \(D^{\mathrm{geo}}_0\) denotes simulator-consistent depth. Invalid or
non-finite pixels are excluded from the fit. This alignment establishes the
internal scene coordinates up to the scale supported by the single-image
hypothesis.

The source image and its initial geometric record are stored permanently:
\begin{equation}
    \mathcal R=
    (I_0,B_0,D^{\mathrm{geo}}_0,V_0,Q_0,L_0,K_0,T_0).
    \label{eq:app_reference_record}
\end{equation}
Here, \(B_0\) is the cached inpainted background plate;
\(V_0\), \(Q_0\), and \(L_0\) denote validity, confidence, and object
labels, while \(K_0\) and \(T_0\) are camera intrinsics and world-to-camera
extrinsics. The
reference image and its geometry remain read-only during online interaction.

\subsection{Declarative Configuration and Physical Simulation}
\label{app:physical_simulation}

\paragraph{Configuration proposal, validation, and override.}
TourPhysics accepts a directly authored physical declaration. As an optional
initializer, Qwen2.5-VL-72B-Instruct~\cite{Qwen2VL} receives the complete image
and the masked object crops and proposes a structured configuration. For object
\(i\), the record contains a material family \(m_i\), continuous parameters
\(\theta_i\), a boundary condition \(b_i\), and an optional initial velocity;
the scene record additionally contains velocity keyframes, scheduled force
fields, and action mappings:
\begin{equation}
    \Theta_{\mathrm{phys}}=
    \left(
    \{m_i,\theta_i,b_i,v_i^0\}_{i=1}^{N},
    \mathcal U,\mathcal F
    \right).
    \label{eq:app_physics_configuration}
\end{equation}
The supported families cover rigid bodies; MPM elastic, elastoplastic,
granular, snow, and liquid materials; and PBD elastic, cloth, liquid, and
particle materials. Their parameters include density, friction, Young's
modulus, Poisson ratio, yield stress or angle, compliance, relaxation, and air
resistance as applicable. Boundary conditions include fixed objects, ground
support, and particle pinning; for example, a declared top fraction of a cloth
mesh can be pinned to represent a hanging attachment. Supported external fields
include constant, wind, point, drag, vortex, turbulence, and noise fields, each
with an activation schedule.

The proposal is parsed against a material and force-field whitelist, completed
with family-specific defaults, and exposed for manual editing. The resulting
declaration is frozen before the first trajectory simulation, so the VLM
participates only in initialization. This frozen record is
\(\Theta_{\mathrm{decl}}\) in the main
paper.

\paragraph{Multi-physics solver routing.}
We use the GPU-accelerated Genesis engine~\cite{Genesis}. Non-deformable objects
and articulated agents are routed to rigid-body dynamics (RBD). In generalized
coordinates, the update solves~\cite{featherstone2014rigid}
\begin{equation}
    M(q)\ddot q+h(q,\dot q)
    =\tau+J(q)^{\mathsf T}f_{\mathrm{ext}},
    \label{eq:app_rbd}
\end{equation}
where \(M\) is the inertia matrix, \(h\) collects velocity-dependent and
gravitational terms, and \(J^{\mathsf T}f_{\mathrm{ext}}\) maps contact and
external forces into generalized coordinates.

Continuum materials and large volumetric deformations are handled with the
Material Point Method (MPM)~\cite{jiang2016material}. Material state is carried
by Lagrangian particles and momentum is updated on an Eulerian grid. In
particular, the internal force accumulated at grid node \(j\) is
\begin{equation}
    f_j^{\mathrm{int}}
    =-\sum_p V_p\sigma_p\nabla w_{jp},
    \label{eq:app_mpm}
\end{equation}
where \(V_p\) and \(\sigma_p\) are particle volume and stress, and \(w_{jp}\) is
the particle--grid interpolation kernel. Particle-to-grid transfer, the grid
momentum update, and grid-to-particle transfer naturally support elastic,
plastic, granular, and fluid trajectories.

Thin shells and constraint-dominated particle systems are handled with
Position-Based Dynamics (PBD)~\cite{muller2007pbd}. For a constraint
\(C(x)=0\), the position correction for particle \(i\) is
\begin{equation}
    \Delta x_i=
    -w_i\frac{C(x)}
    {\sum_j w_j\|\nabla_{x_j}C(x)\|_2^2+\epsilon}
    \nabla_{x_i}C(x),
    \label{eq:app_pbd}
\end{equation}
where \(w_i\) is inverse mass. Iterative projection enforces stretch, bending,
volume, density, and attachment constraints.

RBD--MPM and RBD--PBD contacts are enabled through the engine coupler. The
solvers advance within the same scene step, exchange contact response at their
interfaces, and retain solver-native state for the next committed window.
Static support uses a rigid plane placed at the aligned scene floor. Unless a
scene declaration overrides them, the solver step is \(4\,\mathrm{ms}\), with
10 substeps for rigid-only scenes and 16 for scenes containing particle
materials; gravity is \((0,0,-9.8)\). The default MPM and PBD domains are
\([-3,3]^3\); MPM uses particle size \(0.02\) and grid density \(32\), whereas
PBD uses particle size \(0.01\). These values define the controllable scene
hypothesis.

\paragraph{Authoritative trajectory simulation and observation sampling.}
For every online action \(a_n=(c_n,u_n)\), the simulator and camera controller
are advanced on a working copy of the preceding committed checkpoint before
observation synthesis:
\begin{equation}
    (\mathbf X_n,\mathbf C_n,\mathcal G_n)
    =
    \mathcal F_{\mathrm{seq}}
    (\widetilde X_n,\widetilde C_n,a_n;\Theta_{\mathrm{decl}}),
    \label{eq:app_simulation_sequence}
\end{equation}
The resulting trajectory segment records the simulator state, camera pose,
simulator-consistent depth, visibility, object labels, and geometric confidence
at every observation frame. This segment remains fixed during video generation,
and its terminal state is committed only after the resulting observation passes
the quality gate. At each source frame, the simulator renderer exports metric
depth, entity-ID segmentation, camera intrinsics and extrinsics, and the
dynamic-object state used to derive visibility and motion controls. These
rendered buffers populate \(\mathcal G_n\) and condition observation synthesis.
The video model's decoder supplies the output RGB.

The simulator control record is indexed at \(f_{\mathrm{sim}}=60\) fps, while
generated observations use \(f_{\mathrm{gen}}=16\) fps. Let \(o_n\) be the
sampling origin associated with window \(n\), rather than the committed terminal
state index. To avoid floating-point timestamp drift,
generated frame \(j\) selects simulator frame
\begin{equation}
    k_n(j)=o_n+\left\lfloor
    \frac{(2j+1)f_{\mathrm{sim}}}{2f_{\mathrm{gen}}}
    \right\rfloor.
    \label{eq:app_timeline_mapping}
\end{equation}
The initial origin \(o_0\) follows the simulator's frame-index convention. Set
\(o_{n+1}=o_n+300\), so \(k_{n+1}(0)=k_n(80)\). Thus, under this
center-of-bin convention, the shared boundary is the sampled frame \(k_n(80)\);
the committed terminal states \(X_n^+,C_n^+\) are defined at that endpoint, not
at the raw origin \(o_n+300\). Consecutive observations contain 81 frames and
share one simulator/conditioning boundary, corresponding to an 80-frame stride;
their decoded RGB boundary is checked by the overlap gate and deduplicated when
windows are concatenated. The next action becomes available only after the
current window is committed. The simulated trajectory,
including all intervention and force-field activation indices, remains unchanged during
generation and retry.

\subsection{Separated Depth Controls}
\label{app:depth_controls}

Let \(n\) index committed interaction windows and \(t\in\mathcal W_n\) index
frames within a window; retry attempts retain the same \(n\). TourPhysics
maintains the following per-frame quantities:
\begin{equation}
    D_{n,t}^{\mathrm{geo}},\qquad
    D_{n,t}^{\mathrm{gen}}
    =[\mathcal D_{\mathcal W_n}^{\mathrm{gen}}]_t,\qquad
    D_{n,t}^{\mathrm{base}}
    =[\mathcal D_{\mathcal W_n}^{\mathrm{base}}]_t.
    \label{eq:app_two_depths}
\end{equation}
Simulator-consistent depth \(D_{n,t}^{\mathrm{geo}}\) is metric camera-space
\(z\)-depth in calibrated scene units; \(T_t\) is world-to-camera and
\(\pi^{-1}(p,z;K_t)\) returns a camera-frame point with third coordinate \(z\).
It is used for projection, visibility, object correspondence, and evaluation.
Scene initialization and simulator rendering are its only writers. The base
window tensor
\(\mathcal D_{\mathcal W_n}^{\mathrm{base}}
=\mathcal F_{\mathrm{base}}(\mathcal G_n;\mathcal R)\) is a normalized
relative-depth condition supplied to the video model.

The persistent state stores accepted tail evidence \(\mathcal E_n\), not a
writable depth map. At the beginning of the current window, after
\(\mathcal G_n\) has been fixed, the generator-facing tensor is constructed as
\begin{equation}
    \mathcal D_{\mathcal W_n}^{\mathrm{gen}}
    =\mathcal F_{\mathrm{ctrl}}(\mathcal G_n,\mathcal E_n)
    =(1-W_n)\odot\mathcal D_{\mathcal W_n}^{\mathrm{base}}
    +W_n\odot\overline{\mathcal D}_{\mathcal W_n}(\mathcal E_n),
    \label{eq:app_depth_completion}
\end{equation}
where \(W_n\in[0,1]^{|\mathcal W_n|\times H\times W}\) selects eligible newly
exposed regions inadequately covered by the current generator-facing base
control. The candidate completion is estimated only from committed tail
evidence and is applied only when novelty, overlap, temporal consistency, and
the affine-aligned depth-agreement tests are satisfied. The reported gates are trusted
fraction \(\geq0.20\), Spearman \(\geq0.70\), model-space NRMSE \(\leq0.08\),
and metric-space NRMSE \(\leq0.12\), with novelty, target-hole, and
source-support fractions \(0.05,0.01,0.20\), respectively. These tests are
pointwise after the frame-level admissibility checks; otherwise
\(W_n(t,p)=0\) and the base control is retained exactly at that location.

Generated depth affects only visual conditioning; collision geometry, object
state, and simulator-consistent depth remain unchanged.

\subsection{Observation Generation and Transactional State}
\label{app:observation_commit}

The observation model is a finite-window video diffusion model trained for geometry-conditioned observation synthesis:
\begin{equation}
 Y_{\mathcal W_n}\sim p_\theta\!\left(
 Y_{\mathcal W_n}\mid
 \mathcal G_n,\mathcal D_{\mathcal W_n}^{\mathrm{gen}},I_0,\mathcal K_n
 \right).
 \label{eq:app_observation}
\end{equation}
The physical and camera state sequence and its geometric controls remain fixed
while \(Y_{\mathcal W_n}\) is generated. The native temporal cache is used for
short-range continuity, whereas the appearance memory provides sparse
long-range access to previously accepted static appearance.

For a configured tail subset \(\mathcal T_n\) (the reported profile uses
\(\{64,68,72,76,80\}\)), the candidate evidence is
\begin{equation}
    \widehat{\mathcal E}_{n+1}
    =\Phi_{\mathcal E}\!\left(
    Y_{\mathcal W_n}|_{\mathcal T_n},\mathcal G_n,\mathcal R\right),
    \label{eq:app_staged_evidence}
\end{equation}
and remains private until the quality decision accepts the full window. The
operator stores only generated-depth estimates, trusted/static support,
indices, and alignment/confidence statistics; simulator-consistent geometry
is never overwritten.

The corresponding appearance pages are staged from clean attention features
and reliable static support:
\begin{equation}
    \widehat{\mathcal K}_{n+1}
    =\mathcal K_n\oplus
    \Phi_{\mathrm{page}}\!\left(
    Z_{\mathcal W_n}^{\mathrm{attn}},\mathcal G_n,
    \Omega_{\mathcal W_n}\right),
    \label{eq:app_staged_page_update}
\end{equation}
where \(\Omega_{\mathcal W_n}=\{\Omega_t:t\in\mathcal W_n\}\) is the
reliable static support defined below. The pages remain private, together
with the candidate evidence, until the enclosing window commits.

Candidate updates remain private during generation. If the quality decision is
\(q_n=1\), the state is committed as
\begin{equation}
    \mathcal S_{n+1}
    =
    \left(
    X_n^+,C_n^+,
    \widehat{\mathcal K}_{n+1},
    \widehat{\mathcal E}_{n+1}
    \right).
    \label{eq:app_commit}
\end{equation}
If \(q_n=0\), the persistent state remains unchanged and the working simulator
and camera copy is discarded:
\begin{equation}
    \mathcal S_{n+1}=\mathcal S_n.
    \label{eq:app_reject}
\end{equation}
In this rejection equation the \(n+1\) label is only a bookkeeping view; the
committed index does not advance. The simulator state, memory pages, and depth evidence remain at
the preceding committed boundary; a retry keeps the same committed index and
uses the same action, trajectory, controls, and committed memory.
If \(q_n=1\), \(\widehat{\mathcal E}_{n+1}\) becomes the committed
\(\mathcal E_{n+1}\). After the next action fixes \(\mathcal G_{n+1}\),
Eq.~\eqref{eq:app_depth_completion} constructs
\(\mathcal D_{\mathcal W_{n+1}}^{\mathrm{gen}}\); no candidate evidence can be
used before that commit.

\subsection{Quality Gate, Atomicity, and Retry Policy}
\label{app:quality_gate}

The quality gate is a fail-closed publication check, not an evaluation metric
for physical fidelity. It runs after all 81 candidate RGB frames have been
rendered and decoded, but before the candidate cache state, appearance pages,
tail evidence, or terminal simulator state becomes persistent. The media
contract requires exactly 81 decodable frames at \(832\times480\). A contract
violation rejects the candidate before any state publication.

Let \(Y_0,\ldots,Y_{80}\) be the decoded candidate frames on the 8-bit RGB
scale, and let \(B_n\) be the last frame of the preceding committed window. We
use the full-frame mean absolute RGB difference
\begin{equation}
    \delta(A,B)=\frac{1}{3HW}
    \sum_{p,c}|A(p,c)-B(p,c)|
    \label{eq:app_gate_delta}
\end{equation}
and adjacent-frame changes \(d_t=\delta(Y_t,Y_{t-1})\), \(t=1,\ldots,80\).
For windows after the first, the shared-boundary error and boundary-motion
ratio are
\begin{equation}
    e_{\mathrm{ov}}=\delta(Y_0,B_n),\qquad
    r_{\mathrm{bd}}=\frac{e_{\mathrm{ov}}}
    {\max\{\operatorname{median}(d_1,\ldots,d_8),10^{-6}\}}.
    \label{eq:app_gate_boundary}
\end{equation}
The first window has no preceding boundary and sets both quantities to zero.
We also compute
\begin{align}
    r_{\mathrm{black}}
    &=\frac{1}{81HW}\sum_{t,p}
      \mathbf 1\!\left[\max_cY_t(p,c)\leq3\right],\\
    r_{\mathrm{freeze}}
    &=\frac{1}{80}\sum_{t=1}^{80}\mathbf 1[d_t<0.2].
    \label{eq:app_gate_collapse}
\end{align}
Thus, black pixels and frozen transitions are defined directly on decoded
8-bit frames rather than by a learned quality model.

For the reported runs, the binary decision is the conjunction
\begin{equation}
\begin{split}
q_n={}&\mathbf 1[\text{media contract}]
\mathbf 1[e_{\mathrm{ov}}\leq4.0]
\mathbf 1[r_{\mathrm{bd}}\leq3.0]\\
&\cdot\mathbf 1[r_{\mathrm{black}}\leq0.10]
\mathbf 1[r_{\mathrm{freeze}}\leq0.25]
\chi_{\mathrm{repr}},
\end{split}
\label{eq:app_quality_gate}
\end{equation}
where
\(\chi_{\mathrm{repr}}=\mathbf 1[e_{\mathrm{repr}}\leq24.0]\) when a
pre-commit feedback-reprojection manifest is available, and
\(\chi_{\mathrm{repr}}=1\) otherwise. No manifest was supplied in the reported
video runs, so the optional field was recorded as zero and did not vary their
acceptance decisions. Known-background and dynamic-foreground RGB MAE against
the paired simulator render are also logged on the 0--255 scale. They are
diagnostics only: simulator RGB is a conditioning reference rather than a
pixel target, so these two MAEs do not enter Eq.~\eqref{eq:app_quality_gate}.
Thresholds are fixed for an attempt and are never relaxed after a failure.

If \(q_n=0\), the staged video/cache state and candidate evidence are discarded,
the working simulator/camera copy is released, and the rejection rule together
with the main-paper knowledge-transition rule leave every persistent component
at the last committed boundary. The implementation performs no automatic retry. An
explicit retry starts from the same checkpoint with the same committed index,
action, trajectory, controls, and committed memory, although the caller may
supply a different sampling seed. A later user action becomes available after
a candidate passes the unchanged gate and commits.

\subsection{Reference-Anchored Appearance Memory}
\label{app:appearance_memory_implementation}

Only reliable static regions can enter long-term memory. The eligible support
is defined as
\begin{equation}
    \Omega_t=
    V_t\cap M_t^{\mathrm{sta}}
    \cap\{Q_t\geq\tau_q\}\cap\neg \Gamma_t^{\mathrm{edge}},
    \label{eq:app_memory_support}
\end{equation}
where \(M_t^{\mathrm{sta}}\) is derived from the declared static support and
the motion/deformation field, \(\Gamma_t^{\mathrm{edge}}\) excludes unreliable
boundaries, and \(\tau_q\) is a fixed confidence threshold. Moving and
deforming objects are excluded from long-term memory.
The \(K_t\) and \(T_t\) symbols below denote the camera-intrinsic and
world-to-camera fields of the corresponding geometric record.
The page extraction and private publication rule follow
Eq.~\eqref{eq:app_staged_page_update}; no page is readable before the enclosing
window commits.

For a target pixel \(p\), simulator-consistent depth back-projects the pixel
into the shared scene coordinate system as metric camera-space \(z\)-depth,
with \(T_t\) world-to-camera:
\begin{equation}
    x_w=
    T_t^{-1}
    \pi^{-1}(p,D_t^{\mathrm{geo}}(p);K_t).
    \label{eq:app_backprojection}
\end{equation}
Projection into an earlier committed view \(s\) gives
\begin{equation}
    (\hat p_s,\hat z_s)
    =
    \pi(T_sx_w;K_s).
    \label{eq:app_projection}
\end{equation}
A source view is eligible only when the projected location is valid, visible,
static, and depth-consistent. The correspondence confidence is
\begin{equation}
    c_s(p)=
    Q_t(p)Q_s(\hat p_s)
    \exp\left(
    -\frac{
    |D_s^{\mathrm{geo}}(\hat p_s)-\hat z_s|
    }{\epsilon_s(p)}
    \right).
    \label{eq:app_correspondence}
\end{equation}
Here \(\epsilon_s(p)>0\) is the metric-depth uncertainty tolerance, and the
binary correspondence mask \(m_s(p)\) is one exactly when the projected source
passes the in-bounds, visibility, static-support, and depth-consistency tests.

Let \(A^0\) denote the native reference-conditioned attention output and
\(A^{\mathrm{hist}}\) the output after adding an eligible historical page.
Historical appearance is introduced only as a bounded residual:
\begin{equation}
    A(p)=
    A^0(p)+g_s(p)
    \left(A^{\mathrm{hist}}(p)-A^0(p)\right),
    \label{eq:app_memory_residual}
\end{equation}
where
\begin{equation}
    g_s(p)=
    \lambda_{\mathrm{mem}}m_s(p)
    \operatorname{clip}(c_s(p),0,1).
    \label{eq:app_memory_gate}
\end{equation}
If no valid source view exists, \(m_s(p)=0\), and the output exactly recovers
the native path \(A(p)=A^0(p)\).

A candidate memory page is stored in a private staging area while its
observation is evaluated. It becomes readable only after the observation
window commits. This rule prevents failed or rejected generations from
contaminating future observations.

\section{Additional Evaluation Protocol}
\label{app:evaluation_protocol}

\subsection{Benchmark and Comparison Setting}

The evaluation contains simulator-defined camera tours and object
manipulations involving rigid, cloth, elastic, granular, and fluid proxies.
The benchmark is initialized from 61 source-image identities and contains 108
interaction sequences. When several sequences share an input image, we report
both sequence-level and identity-balanced statistics where appropriate.

The benchmark evaluates agreement with a declared simulation trajectory. It does
not test whether the chosen declarative parameters are the true physical
parameters of the source scene. This distinction is important because
single-image initialization does not uniquely determine scale, occluded
geometry, density, friction, or deformation parameters.

Compared methods expose different conditioning interfaces. Some methods
receive only an image, while others receive an explicit camera trajectory,
event description, or privileged control. Such differences are indicated in
the corresponding tables. Methods producing shorter videos may be temporally
normalized for generic video-quality diagnostics, but these results are not
treated as equivalent to native online generation.

\subsection{Camera and Object Motion}

\label{app:camera_object_protocol}

\paragraph{Camera coordinates and alignment.}
We retain only the 49 sequences whose reference poses are exact runtime
exports. For sequence \(s\), ViPE recovers camera centers
\(\widehat c_{s,t}\), which are timestamp-matched to reference centers
\(c_{s,t}\) in the internal simulator world coordinate system. One Umeyama
similarity transform \((\gamma_s,R_s,u_s)\) is fitted over every valid frame of
the complete sequence, never separately by window or segment:
\begin{equation}
    \widetilde c_{s,t}=\gamma_sR_s\widehat c_{s,t}+u_s.
    \label{eq:app_camera_sim3}
\end{equation}
If \(\mathcal V_s\) is the valid timestamp set, the per-sequence absolute
trajectory error (ATE) is
\begin{equation}
    \mathrm{ATE}_s=
    \sqrt{\frac{1}{|\mathcal V_s|}
    \sum_{t\in\mathcal V_s}
    \|\widetilde c_{s,t}-c_{s,t}\|_2^2}.
    \label{eq:app_camera_ate}
\end{equation}
ATE is expressed in simulator world units after alignment; those units belong
to the declared scene hypothesis and are not recovered real-world meters. The
path-length ratio (PLR) uses the same valid frames and transform:
\begin{equation}
    \mathrm{PLR}_s=
    \frac{\sum_t\|\widetilde c_{s,t}-\widetilde c_{s,t-1}\|_2}
    {\sum_t\|c_{s,t}-c_{s,t-1}\|_2}.
    \label{eq:app_camera_plr}
\end{equation}
It is dimensionless and has ideal value one because the recovered and
reference cameras then travel equal total distance. The main-paper
camera/object-motion table reports equal-weight means over valid sequences for
both ATE and PLR.

T-Dir and R-Dir use schedule segments that are fully covered, last at least
0.5\,s, and contain exclusively translation or rotation commands,
respectively. For each segment we form the recovered and reference relative
transforms between its endpoints. Let \(v^{\mathrm{pred}}\) and
\(v^{\mathrm{ref}}\) denote their translation vectors for T-Dir or their
axis--angle rotation vectors for R-Dir. A direction hit is
\begin{equation}
    h=\mathbf 1\!\left[
    \frac{v^{\mathrm{pred}}\!\cdot v^{\mathrm{ref}}}
    {\|v^{\mathrm{pred}}\|_2\|v^{\mathrm{ref}}\|_2}>0\right].
    \label{eq:app_camera_direction}
\end{equation}
The test therefore checks the correct half-space, not exact-axis recovery.
Hits are averaged first over eligible segments in a sequence and then equally
over sequences; the table reports percentages.

\paragraph{Object coordinates and aggregation.}
The paired simulator and generated videos are sampled at 8\,fps and
\(682\times384\). Frozen initial object masks seed the same CoTracker3
multi-point procedure in both videos, and a robust center is computed for each
dynamic object. With \(D=\sqrt{682^2+384^2}\), reference/generated centers
\(x_{s,i,t},\widehat x_{s,i,t}\), and joint-visibility indicator \(v_{s,i,t}\),
the normalized error is
\begin{equation}
    e_{s,i,t}=\frac{\|\widehat x_{s,i,t}-x_{s,i,t}\|_2}{D}.
    \label{eq:app_object_error}
\end{equation}
Thus, the 0.05 threshold is five percent of the image diagonal in normalized
image coordinates. We retain a dynamic object only if at least eight frames
have jointly visible robust centers and its joint-visibility fraction is at
least 0.15. For a retained object, with
\(N_{s,i}=\sum_tv_{s,i,t}\) and last jointly visible frame \(t^*_{s,i}\),
\begin{align}
    \mathrm{ADE}_{s,i}&=\frac{1}{N_{s,i}}
       \sum_t v_{s,i,t}e_{s,i,t},\\
    \mathrm{End}_{s,i}&=e_{s,i,t^*_{s,i}},\\
    \mathrm{S@0.05}_{s,i}&=\mathbf 1[\mathrm{ADE}_{s,i}\leq0.05],\\
    \mathrm{Vis}_{s,i}&=N_{s,i}/T_s.
    \label{eq:app_object_metrics}
\end{align}
Each quantity is averaged over retained objects within a sequence and then
equally over sequences. ADE, End Err., S@0.05, and Vis. are dimensionless and
share the same method-specific support.

\paragraph{Effective support.}
Table~\ref{tab:app_camera_object_support} reports the method-specific support
after visibility filtering. The table note states the unit and ordering of each
count; identities are distinct input-image SHA-256 values.

\begin{table}[!t]
    \caption{Effective support for the camera- and object-motion protocols. Each
    cell lists counts in the order specified in the table note; ``--'' denotes an
    incompatible protocol.}
    \label{tab:app_camera_object_support}
    \begin{center}
        \setlength{\tabcolsep}{7.0pt}
        \begin{tabularx}{\linewidth}{@{}l *{4}{Y}@{}}
                \toprule
                \multirow{2}{*}{Method}
                & \multicolumn{3}{c}{Camera}
                & \multicolumn{1}{c}{Object Motion} \\
                \cmidrule(lr){2-4}\cmidrule(lr){5-5}
                & ATE/PLR & T-Dir & R-Dir & Object Motion \\
                \midrule
                LingBot-Cam & 49/40 & 70/48/40 & 52/25/25 & 74/49/27 \\
                LingBot-Act & 49/40 & 70/48/40 & 52/25/25 & 97/51/29 \\
                Wan2.1-I2V & 49/40 & 70/48/40 & 52/25/25 & 165/92/49 \\
                Sora-2 & 49/40 & 57/45/37 & 1/1/1 & 186/97/53 \\
                MotionCtrl & 49/40 & 70/48/40 & 52/25/25 & 179/95/51 \\
                Tora-Initiator & -- & -- & -- & 175/96/52 \\
                minWM-Wan & 49/40 & 70/48/40 & 52/25/25 & 177/95/51 \\
                \midrule
                TourPhysics & 49/40 & 70/48/40 & 52/25/25 & 168/93/49 \\
                \bottomrule
        \end{tabularx}
        \par\vspace{3pt}
        \parbox{0.98\linewidth}{ATE/PLR counts are sequences/identities;
        T-Dir and R-Dir counts are segments/sequences/identities; object-motion
        counts are objects/sequences/identities. All counts are after
        method-specific visibility filtering.}
    \end{center}
\end{table}

\subsection{Simulation-Trajectory Adherence}
\label{app:physics_protocol}

The six deterministic metrics use the same frozen-mask, multi-point CoTracker3
observations from paired simulator/generated RGB at 8\,fps and
\(682\times384\). All distances below are divided by the image diagonal
\(D=\sqrt{682^2+384^2}\). A record is first reduced within its sequence; the
reported value is then an equal-weight mean over sequences. Consequently,
sequences with several objects or events do not receive extra weight.

\paragraph{Contact Err.}
For initiator point set \(P_k\) and target point set \(Q_k\), the separation
curve is
\begin{equation}
    g_k=\min_{p\in P_k,q\in Q_k}\|p-q\|_2/D.
    \label{eq:app_contact_curve}
\end{equation}
We median-filter \(g_k\) over a three-frame neighborhood and search from the
declared onset through the next 8\,s. At least eight valid samples and an
approach amplitude of 0.002 diagonal are required. The initial level is the
maximum in the early part of the search window (up to 16 samples), and the
minimum level is its fifth percentile. The contact proxy is the first sample
below the minimum plus 15\% of the initial-to-minimum range, or the closest
approach if no sample crosses that level. The simulator selects the earliest
valid target, and the generated video is scored against that same target. If
the generated target is observable but has no valid approach, its error is the
8\,s search-window penalty. Evaluation frame \(k\) is converted by
\(t_k=k/8\) seconds (equivalently, by its stored 8-fps timestamp), and the
reported record is \(|t_k^{\mathrm{gen}}-t_k^{\mathrm{sim}}|\) in seconds.
Full-duration Wan and MotionCtrl tracks use their timestamp-mapped evaluation
timeline.

\paragraph{Response Acc.}
For each simulator-responsive target, a 2D linear-velocity model is fitted to
its robust center from 1.0 to 0.125\,s before contact. Response magnitude is
the median normalized residual from that extrapolation between 1 and 2\,s
after contact. A reference target is applicable only if its residual is at
least 0.006. The generated response is correct when its residual is at least
\(\max(0.006,0.5r_{\mathrm{sim}})\). An observable target with no generated
approach is a failure; insufficient visibility is excluded. Response Acc. is
the fraction of correct applicable targets, expressed as a percentage.

\paragraph{Overlap Exc.}
For two tracked point sets, let
\begin{equation}
    o_k=\frac{\operatorname{area}
    (\operatorname{hull}(P_k)\cap\operatorname{hull}(Q_k))}
    {\min\{\operatorname{area}(\operatorname{hull}(P_k)),
           \operatorname{area}(\operatorname{hull}(Q_k))\}}.
    \label{eq:app_overlap}
\end{equation}
The metric is
\(\max\{0,Q_{0.9}(o^{\mathrm{gen}})-Q_{0.9}(o^{\mathrm{sim}})\}\)
over the 2\,s following each sequence's visual contact. It is a dimensionless
excess 2D hull-overlap fraction; zero is not evidence of zero 3D penetration.

\paragraph{Shape Err. and Area Drift.}
For each deformable object and frame with at least six jointly tracked points,
the simulator and generated point configurations are independently centered
and divided by their root-mean-square radius. The generated configuration is
then aligned to the simulator configuration by the optimal 2D Procrustes
rotation. Shape Err. is the mean corresponding-point distance after this
normalization and is dimensionless. An object record requires at least eight
valid frames and averages their errors.

On the same jointly visible points, let
\(a_k=A_k^{\mathrm{gen}}/A_k^{\mathrm{sim}}\), where each \(A_k\) is the
tracked-point convex-hull area. With
\(a_0\) defined as the median of the first \(\min(10,N)\) valid ratios, Area
Drift is
\begin{equation}
    \frac{1}{N}\sum_{k=1}^{N}|a_k/a_0-1|.
    \label{eq:app_area_drift}
\end{equation}
It is a dimensionless projected-area proxy, not a mass-conservation score, and
also requires at least eight valid frames.

\paragraph{Recovery Acc.}
For each declared deformation interval, shape distance is measured relative
to the last valid pre-event frame using the same Procrustes normalization.
Let \(p\) be its 90th percentile during the event and \(r\) its 20th
percentile from 0.5 to 2.5\,s after the event. The recovery fraction is
\(\rho=\operatorname{clip}((p-r)/p,-1,1)\). An event is applicable when
\(p_{\mathrm{sim}}\geq0.015\), \(\rho_{\mathrm{sim}}>0.05\), and both videos
have valid measurements. It is correct when
\begin{equation}
    p_{\mathrm{gen}}\geq0.5p_{\mathrm{sim}},\qquad
    \rho_{\mathrm{gen}}\geq
    \max\{0.10,\rho_{\mathrm{sim}}-0.30\}.
    \label{eq:app_recovery}
\end{equation}
Recovery Acc. is the event accuracy, averaged event-first within sequence and
then across sequences.

VideoPhy-2 AutoEval is separate from these paired metrics. SA expands to
Semantic Adherence and PC to Physical Commonsense; both are custom-data
VideoPhy-2 means on a 1--5 scale using the frozen shared prompts. Avg is the
arithmetic mean \((\mathrm{SA}+\mathrm{PC})/2\), not an official VideoPhy-2
aggregate. These scores are auxiliary perceptual diagnostics and do not verify
the declared trajectory.

\paragraph{Effective support.}
Table~\ref{tab:app_physics_support} lists the method-specific supports for the
deterministic metrics and the auxiliary VideoPhy-2 scores. For the first six
metric columns, each cell is ordered as records/sequences/identities; the final
three columns are ordered as sequences/identities. The counts are taken from the
existing records, and identities are distinct input-image SHA-256 values.

\begin{table}[!t]
    \caption{Effective support for simulator-trajectory adherence and auxiliary
    VideoPhy-2 diagnostics. ``--'' denotes an incompatible protocol.}
    \label{tab:app_physics_support}
    \begin{center}
        \setlength{\tabcolsep}{3.0pt}
        \begin{tabularx}{\linewidth}{@{}l *{9}{Y}@{}}
                \toprule
                \multirow{2}{*}{Method}
                & \multicolumn{3}{c}{Collision Proxies}
                & \multicolumn{3}{c}{Deformation Proxies}
                & \multicolumn{3}{c}{\shortstack{VideoPhy-2 AutoEval\\(Auxiliary)}} \\
                \cmidrule(lr){2-4}\cmidrule(lr){5-7}\cmidrule(lr){8-10}
                & Contact Err. & Response Acc. & Overlap Exc.
                & Shape Err. & Area Drift & Recovery Acc.
                & SA & PC & Avg \\
                \midrule
                LingBot-Cam & 14/14/6 & 1/1/1 & 2/2/2 & 19/17/8 & 19/17/8 & 6/2/1 & 108/61 & 108/61 & 108/61 \\
                LingBot-Act & 14/14/6 & 2/2/2 & 1/1/1 & 27/24/12 & 27/24/12 & 3/3/2 & 108/61 & 108/61 & 108/61 \\
                Wan2.1-I2V & 14/14/6 & 3/3/2 & 6/6/5 & 31/26/14 & 31/26/14 & 11/5/2 & 108/61 & 108/61 & 108/61 \\
                Sora-2 & 14/14/6 & 7/7/3 & 8/8/4 & 31/26/14 & 31/26/14 & 4/3/2 & 108/61 & 108/61 & 108/61 \\
                MotionCtrl & 14/14/6 & 8/8/3 & 4/4/3 & 31/26/14 & 31/26/14 & -- & 108/61 & 108/61 & 108/61 \\
                Tora-Initiator & 14/14/5 & 3/3/1 & 4/4/2 & 29/24/13 & 29/24/13 & 1/1/1 & 97/53 & 97/53 & 97/53 \\
                minWM-Wan & 14/14/6 & 8/8/3 & 9/9/4 & 31/26/14 & 31/26/14 & 12/9/3 & 108/61 & 108/61 & 108/61 \\
                \midrule
                TourPhysics & 14/14/6 & 5/5/3 & 9/9/5 & 30/25/13 & 30/25/13 & 8/7/2 & 108/61 & 108/61 & 108/61 \\
                \bottomrule
        \end{tabularx}
        \par\vspace{3pt}
        \parbox{0.98\linewidth}{The Contact, Response, Overlap, Shape,
        Area, and Recovery columns list records/sequences/identities. The SA, PC,
        and Avg columns list sequences/identities. Counts are after
        method-specific support filtering; ``--'' denotes an incompatible
        protocol.}
    \end{center}
\end{table}

\subsection{Appearance and Long-Horizon Consistency}
\label{app:appearance_protocol}

We evaluate input-scene subject consistency, background consistency, temporal
stability, and camera-motion classification accuracy. TourPhysics obtains
input-video subject consistency of \(0.9743\), input-video background
consistency of \(0.9791\), and camera-motion classification accuracy of
\(0.8921\) under the reported protocol.

\paragraph{Cache drift.}
For a decoded RGB video \(I_0,\ldots,I_{T-1}\), define the mean-channel
luminance proxy
\begin{equation}
    L_t=\frac{1}{3HW}\sum_{p,c}I_t(p,c).
    \label{eq:app_luminance}
\end{equation}
It is the arithmetic mean of decoded 8-bit channels, not a gamma-corrected
photometric luminance. Let \(m=\max\{1,T-\lfloor0.8T\rfloor\}\). The two
drift measures in the main-paper memory-diagnostics table are
\begin{align}
    D_{\mathrm{end}}&=|L_{T-1}-L_0|,\\
    D_{\mathrm{tail}}&=
    \left|\frac{1}{m}\sum_{t=T-m}^{T-1}L_t
    -\frac{1}{m}\sum_{t=0}^{m-1}L_t\right|.
    \label{eq:app_cache_drift}
\end{align}
Both are unnormalized 0--255 intensity-level differences; they are not
percentages. The reported entries are equal-weight sequence means over the
same 72 paired ordinary-cache and persistent-sink videos. These 72 sequences
come from 27 distinct input-image identities. The memory router has nonempty
causal revisit targets in 70 of those sequences, covering 26 identities, but
all 72 sequences contribute to the cache-only drift comparison.

\paragraph{Historical-region error.}
The frozen appearance plan associates target latent \(j\) with an earlier
source latent \(h(j)\); their decoded frame indices are \(4j\) and \(4h(j)\).
Both frames are resized to the memory-token lattice, and the earlier frame is
remapped by the plan's source-coordinate field. Let \(M_{j,p}\) indicate a
valid, visible, static, and depth-consistent historical correspondence and let
\(\kappa_{j,p}\in[0,1]\) be its frozen confidence. With \(W_j(\cdot)\) denoting
that remap, the base historical error is the confidence-weighted RGB
Charbonnier error
\begin{equation}
\begin{split}
E_{\mathrm{hist}}={}&
\frac{1}{\sum_{j,p}M_{j,p}\kappa_{j,p}}
\sum_{j,p}M_{j,p}\kappa_{j,p}\\
&\quad\cdot\frac{1}{3}\sum_c
\sqrt{\left(
\frac{I_{4j}(p,c)-W_j(I_{4h(j)})(p,c)}{255}
\right)^2+10^{-6}} .
\end{split}
\label{eq:app_history_error}
\end{equation}
Thus, its region is exactly the routed historical-correspondence mask, and its
base error is dimensionless.

\paragraph{Known-region error.}
Let \(K_{t,p}\) be the simulator-consistent known-background mask
(\texttt{reference\_bg\_lv}, with the frozen preserve/calibration-valid mask
as fallback). It excludes the tracked dynamic foreground. Against the paired
simulator RGB \(I^{\mathrm{sim}}\), the base error is
\begin{equation}
    E_{\mathrm{known}}=
    \frac{\sum_{t,p,c}K_{t,p}
    |I^{\mathrm{gen}}_t(p,c)-I^{\mathrm{sim}}_t(p,c)|}
    {3\sum_{t,p}K_{t,p}},
    \label{eq:app_known_error}
\end{equation}
an MAE in 0--255 intensity levels, averaged over the memory-intervention
chunks. This quantity is a guardrail diagnostic, as specified in the
``Window-level quality gate'' paragraph of the main-paper ``Quality-Gated
Atomic Commit'' subsection; it is not an acceptance threshold.

The appearance-memory treatment and control use the same physical and camera
trajectories, generator-facing depth, diffusion configuration, prompts,
checkpoints, and random seeds. For a scene \(s\), a paired percentage is
\(\Delta_s(E)=100(E_s^{\mathrm{mem}}-E_s^{\mathrm{off}})/
E_s^{\mathrm{off}}\), so negative is favorable. History change is the
equal-weight mean of \(\Delta_s(E_{\mathrm{hist}})\) in both subsets. The
frozen development report also uses the mean of four paired
\(\Delta_s(E_{\mathrm{known}})\) values. The held-out known-region guardrail
uses the relative change of identity-macro MAE,
\(100(\overline E^{\mathrm{mem}}_{\mathrm{known}}-
\overline E^{\mathrm{off}}_{\mathrm{known}})/
\overline E^{\mathrm{off}}_{\mathrm{known}}\), which yields the reported
\(-0.90\%\). Stating these frozen aggregations explicitly avoids conflating a
mean of percentages with a percentage of means.

\paragraph{Memory subsets.}
The four-sequence development stress set uses four distinct input identities
and covers rigid, cloth, and elastic content. Each selected sequence has six
windows (481 output frames), 51 planned target latents, and active revisit
chunks 3--5. Selection favored interpretable revisit/occlusion pressure,
geometric-correspondence support, material diversity, and qualitative
readability; earlier mechanism-screening metadata was available, so this is
not an unbiased generalization set.

The held-out set contains all 14 eligible dynamic input identities unused by
the mechanism pilots or development set. Camera-only identities are excluded,
and one scene per remaining identity is chosen as the lexicographically first
eligible six-window sequence satisfying the same 481-frame, 51-target-latent,
chunks-3--5 contract. This identity-disjoint selection was frozen before
treatment outcomes were inspected. Consequently, the main-paper
memory-diagnostics table reports four development identities and 14 held-out
identities, not repeated variants of a smaller image set.

\section{Additional Ablations and Limitations}
\label{app:additional_ablations}

\subsection{Cache and Memory Ablations}

The persistent-sink cache reduces long-horizon brightness drift compared with
an ordinary finite temporal cache. On the long-video subset, end-to-start drift
decreases from \(26.33\) to \(11.17\), while tail-to-head drift decreases from
\(20.38\) to \(7.74\).

With the observation backbone and generator-facing depth fixed, geometry-routed
appearance memory reduces historical-region error by \(8.25\%\) on the
development split and \(2.39\%\) on the held-out split. Known-region error
decreases by \(2.99\%\) and \(0.90\%\), respectively. The smaller held-out
improvement reflects variation in camera coverage and geometric
correspondence quality.

The fallback property is verified by disabling all valid historical
correspondences. In this case, \(g_s(p)=0\) for all \(p\), and Eq.
\eqref{eq:app_memory_residual} recovers the native observation path exactly.
This test confirms that unavailable or unreliable memory does not introduce an
additional visual bias.

\subsection{Limitations}

TourPhysics constructs a controllable physical hypothesis rather than a unique
real-scene reconstruction. Its physical behavior depends on the supplied
object proxies, material families, boundary conditions, and simulator
parameters. Therefore, adherence to the prescribed simulation trajectory does not guarantee real-world
physical accuracy.

The current geometry is incomplete in occluded or newly exposed regions.
Appearance memory also excludes moving and deforming objects, which limits its
ability to preserve object-specific appearance after large physical changes.
Incorrect camera or depth estimates can further reduce valid
cross-view correspondences.

Finally, finite-window diffusion is not real-time, and uncompressed historical
pages grow with the number of accepted views. Future work should investigate
fixed-capacity page eviction, object-local memory for deforming surfaces,
real-world physical evaluation, and longer interactive tours.


\begin{thebibliography}{99}

\bibitem{zhang2024physdreamer}
T. Zhang, H.-X. Yu, R. Wu, B. Y. Feng, C. Zheng, N. Snavely, J. Wu, and W. T. Freeman, ``PhysDreamer: Physics-Based Interaction with 3D Objects via Video Generation,'' 2024, arXiv:2404.13026.

\bibitem{openai_sora2_2025}
OpenAI, ``Sora 2 is here,'' 2025. [Online]. Available: \url{https://openai.com/index/sora-2/}

\bibitem{muller2007pbd}
M. M{\"u}ller, B. Heidelberger, M. Hennix, and J. Ratcliff, ``Position Based Dynamics,'' \textit{Journal of Visual Communication and Image Representation}, vol. 18, no. 2, pp. 109--118, 2007, doi: 10.1016/j.jvcir.2007.01.005.

\bibitem{croitoru2023diffusion}
F.-A. Croitoru, V. Hondru, R. T. Ionescu, and M. Shah, ``Diffusion Models in Vision: A Survey,'' \textit{IEEE Trans. Pattern Anal. Mach. Intell.}, vol. 45, no. 9, pp. 10850--10869, 2023, doi: 10.1109/TPAMI.2023.3261988.

\bibitem{kong2024hunyuanvideo}
W. Kong, Q. Tian, Z. Zhang, R. Min, Z. Dai, J. Zhou, J. Xiong, X. Li, B. Wu, J. Zhang, K. Wu, Q. Lin et al., ``HunyuanVideo: A Systematic Framework For Large Video Generative Models,'' 2024, arXiv:2412.03603.

\bibitem{gao2025longviemultimodalguidedcontrollableultralong}
J. Gao, Z. Chen, X. Liu, J. Feng, C. Si, Y. Fu, Y. Qiao, and Z. Liu, ``LongVie: Multimodal-Guided Controllable Ultra-Long Video Generation,'' 2025, arXiv:2508.03694. [Online]. Available: \url{https://arxiv.org/abs/2508.03694}.

\bibitem{huang2026cineweaver}
Y. Huang, Y. Chen, W. Dai, Z. Zheng, H. Huang, C. Zhang, J. Zou, H. Xiong, and X. Li, ``CineWeaver: Training-Free Reference-Controllable Multi-Shot Long Video Generation for Cinematic Storytelling,'' 2026, arXiv:2607.26529. [Online]. Available: \url{https://arxiv.org/abs/2607.26529}.

\bibitem{yi2025deepforcing}
J. Yi, W. Jang, P. H. Cho, J. Nam, H. Yoon, and S. Kim, ``Deep Forcing: Training-Free Long Video Generation with Deep Sink and Participative Compression,'' 2025, arXiv:2512.05081. [Online]. Available: \url{https://arxiv.org/abs/2512.05081}.

\bibitem{jiang2022c2matching}
Y. Jiang, K. C. K. Chan, X. Wang, C. C. Loy, and Z. Liu, ``Reference-Based Image and Video Super-Resolution via C$^{2}$-Matching,'' \textit{IEEE Trans. Pattern Anal. Mach. Intell.}, vol. 45, no. 7, pp. 8874--8887, 2023, doi: 10.1109/TPAMI.2022.3231089.

\bibitem{hu2025drsa}
W. Hu, J. T. Hoe, J. Li, H. Hu, X. Jiang, and Y.-P. Tan, ``Cascaded Dynamic Memory Refinement and Semantic Alignment for Exo-to-Ego Cross-View Video Generation,'' \textit{IEEE Trans. Pattern Anal. Mach. Intell.}, vol. 47, no. 9, pp. 7490--7505, 2025, doi: 10.1109/TPAMI.2025.3569195.

\bibitem{bruce2024genie}
J. Bruce, M. Dennis, A. Edwards, J. Parker-Holder, Y. Shi, E. Hughes, M. Lai, A. Mavalankar, R. Steigerwald, C. Apps, Y. Aytar, S. Bechtle et al., ``Genie: Generative Interactive Environments,'' 2024, arXiv:2402.15391.

\bibitem{deepmind2024genie2}
Google DeepMind, ``{Genie}~2: A Large-Scale Foundation World Model,'' 2024. [Online]. Available: \url{https://deepmind.google/discover/blog/genie-2-a-large-scale-foundation-world-model/}

\bibitem{yang2023unisim}
S. Yang, Y. Du, K. Ghasemipour, J. Tompson, L. Kaelbling, D. Schuurmans, and P. Abbeel, ``Learning Interactive Real-World Simulators,'' 2023, arXiv:2310.06114.

\bibitem{decart2024oasis}
Decart, J. Quevedo, Q. McIntyre, S. Campbell, X. Chen, and R. Wachen, ``Oasis: A Universe in a Transformer,'' 2024. [Online]. Available: \url{https://oasis-model.github.io/}

\bibitem{valevski2024gamengen}
D. Valevski, Y. Leviathan, M. Arar, and S. Fruchter, ``Diffusion Models Are Real-Time Game Engines,'' 2025, arXiv:2408.14837.

\bibitem{che2024gamegenx}
H. Che, X. He, Q. Liu, C. Jin, and H. Chen, ``{GameGen-X}: Interactive Open-world Game Video Generation,'' 2024, arXiv:2411.00769.

\bibitem{menapace2022playable}
W. Menapace, S. Lathuili\`ere, A. Siarohin, C. Theobalt, S. Tulyakov, V. Golyanik, and E. Ricci, ``Playable Environments: Video Manipulation in Space and Time,'' in \textit{Proc. IEEE/CVF Conf. Comput. Vis. Pattern Recognit.}, 2022, pp. 3584--3593.

\bibitem{kim2026dexterous}
B. Kim, T. Kim, J. Lee, and H. Joo, ``Dexterous World Models,'' in \textit{Proc. IEEE/CVF Conf. Comput. Vis. Pattern Recognit.}, 2026, pp. 29663--29673.

\bibitem{liu2024physgen}
S. Liu, Z. Ren, S. Gupta, and S. Wang, ``PhysGen: Rigid-Body Physics-Grounded Image-to-Video Generation,'' 2024, arXiv:2409.18964.

\bibitem{chen2025physgen3d}
B. Chen, H. Jiang, S. Liu, S. Gupta, Y. Li, H. Zhao, and S. Wang, ``PhysGen3D: Crafting a Miniature Interactive World from a Single Image,'' in \textit{Proc. IEEE/CVF Conf. Comput. Vis. Pattern Recognit.}, 2025, pp. 6178--6189.

\bibitem{li2024generativedynamics}
Z. Li, R. Tucker, N. Snavely, and A. Holynski, ``Generative Image Dynamics,'' in \textit{Proc. IEEE/CVF Conf. Comput. Vis. Pattern Recognit.}, 2024, pp. 24142--24153.

\bibitem{shen2026phantom}
Y. Shen, J. Xiong, T. Yu, and I. Lourentzou, ``PHANTOM: Physics-Infused Video Generation via Joint Modeling of Visual and Latent Physical Dynamics,'' in \textit{Proc. IEEE/CVF Conf. Comput. Vis. Pattern Recognit.}, 2026, pp. 11185--11194.

\bibitem{ma2026miramo}
X. Ma, Y. Wang, G. Jia, X. Chen, T.-T. Wong, and C. Chen, ``Consistent and Controllable Image Animation With Linear Motion Diffusion Transformers,'' \textit{IEEE Trans. Pattern Anal. Mach. Intell.}, vol. 48, no. 7, pp. 7436--7450, 2026, doi: 10.1109/TPAMI.2026.3664227.

\bibitem{tan2024physmotion}
X. Tan, Y. Jiang, X. Li, Z. Zong, T. Xie, Y. Yang, and C. Jiang, ``PhysMotion: Physics-Grounded Dynamics From a Single Image,'' 2024, arXiv:2411.17189.

\bibitem{wang2025physctrl}
C. Wang, C. Chen, Y. Huang, Z. Dou, Y. Liu, J. Gu, and L. Liu, ``PhysCtrl: Generative Physics for Controllable and Physics-Grounded Video Generation,'' 2025, arXiv:2509.20358.

\bibitem{gillman2025forceprompting}
N. Gillman, C. Herrmann, M. Freeman, D. Aggarwal, E. Luo, D. Sun, and C. Sun, ``Force Prompting: Video Generation Models Can Learn and Generalize Physics-based Control Signals,'' 2025, arXiv:2505.19386.

\bibitem{foo2026physicalsim}
L. G. Foo, M. H. Huang, A. Lattas, S. Moschoglou, T. Beeler, and C. Theobalt, ``Physical Simulator In-the-Loop Video Generation,'' in \textit{Proc. IEEE/CVF Conf. Comput. Vis. Pattern Recognit.}, 2026, pp. 4301--4311.

\bibitem{jiang2016material}
C. Jiang, C. Schroeder, J. Teran, A. Stomakhin, and A. Selle, ``The Material Point Method for Simulating Continuum Materials,'' in \textit{ACM SIGGRAPH 2016 Courses}, 2016, pp. 1--52, doi: 10.1145/2897826.2927348.

\bibitem{wan2025open}
Team Wan, A. Wang, B. Ai, B. Wen, C. Mao, C.-W. Xie, D. Chen, F. Yu, H. Zhao, J. Yang, J. Zeng, J. Wang et al., ``Wan: Open and Advanced Large-Scale Video Generative Models,'' 2025, arXiv:2503.20314.

\bibitem{blattmann2023svd}
A. Blattmann, T. Dockhorn, S. Kulal, D. Mendelevitch, M. Kilian, D. Lorenz, Y. Levi, Z. English, V. Voleti, A. Letts, V. Jampani, and R. Rombach, ``Stable Video Diffusion: Scaling Latent Video Diffusion Models to Large Datasets,'' 2023, arXiv:2311.15127.

\bibitem{karaev2024cotracker3}
N. Karaev, I. Makarov, J. Wang, N. Neverova, A. Vedaldi, and C. Rupprecht, ``CoTracker3: Simpler and Better Point Tracking by Pseudo-Labelling Real Videos,'' 2024, arXiv:2410.11831.

\bibitem{wang2025moge}
R. Wang, S. Xu, C. Dai, J. Xiang, Y. Deng, X. Tong, and J. Yang, ``MoGe: Unlocking Accurate Monocular Geometry Estimation for Open-Domain Images with Optimal Training Supervision,'' in \textit{Proc. IEEE/CVF Conf. Comput. Vis. Pattern Recognit.}, 2025, pp. 5261--5271.

\bibitem{yu2025viewcrafter}
W. Yu, J. Xing, L. Yuan, W. Hu, X. Li, Z. Huang, X. Gao, T.-T. Wong, Y. Shan, and Y. Tian, ``ViewCrafter: Taming Video Diffusion Models for High-Fidelity Novel View Synthesis,'' \textit{IEEE Transactions on Pattern Analysis and Machine Intelligence}, pp. 1--18, 2025, doi: 10.1109/TPAMI.2025.3613256.

\bibitem{chen2025v3d}
Z. Chen, Y. Wang, F. Wang, Z. Wang, F. Sun, and H. Liu, ``V3D: Video Diffusion Models Are Effective 3D Generators,'' \textit{IEEE Trans. Pattern Anal. Mach. Intell.}, pp. 1--18, 2025, doi: 10.1109/TPAMI.2025.3581312.

\bibitem{wang2025video4dgen}
Y. Wang, G. Liu, X. Wang, Z. Chen, J. Li, X. Liang, F. Sun, and J. Zhu, ``Video4DGen: Enhancing Video and 4D Generation Through Mutual Optimization,'' \textit{IEEE Trans. Pattern Anal. Mach. Intell.}, pp. 1--18, 2025, doi: 10.1109/TPAMI.2025.3550031.

\bibitem{cai2024generativerendering}
S. Cai, D. Ceylan, M. Gadelha, C.-H. P. Huang, T. Y. Wang, and G. Wetzstein, ``Generative Rendering: Controllable 4D-Guided Video Generation with 2D Diffusion Models,'' in \textit{Proc. IEEE/CVF Conf. Comput. Vis. Pattern Recognit.}, 2024, pp. 7611--7620.

\bibitem{he2025cameractrlii}
H. He, C. Yang, S. Lin, Y. Xu, M. Wei, L. Gui, Q. Zhao, G. Wetzstein, L. Jiang, and H. Li, ``CameraCtrl II: Dynamic Scene Exploration via Camera-Controlled Video Diffusion Models,'' in \textit{Proc. IEEE/CVF Int. Conf. Comput. Vis.}, 2025, pp. 13416--13426.

\bibitem{ren2025gen3c}
X. Ren, T. Shen, J. Huang, H. Ling, Y. Lu, M. Nimier-David, T. M{\"u}ller, A. Keller, S. Fidler, and J. Gao, ``GEN3C: 3D-Informed World-Consistent Video Generation with Precise Camera Control,'' in \textit{Proc. IEEE/CVF Conf. Comput. Vis. Pattern Recognit.}, 2025, pp. 6121--6132.

\bibitem{zhang2026worldstereo}
Y. Zhang, C. Cao, T. Wang, X. Zuo, J. Wu, J. Zhu, and C. Guo, ``WorldStereo: Bridging Camera-Guided Video Generation and Scene Reconstruction via 3D Geometric Memories,'' in \textit{Proc. IEEE/CVF Conf. Comput. Vis. Pattern Recognit.}, 2026, pp. 40327--40339.

\bibitem{wang2025cut3r}
Q. Wang, Y. Zhang, A. Holynski, A. A. Efros, and A. Kanazawa, ``Continuous 3D Perception Model with Persistent State,'' in \textit{Proc. IEEE/CVF Conf. Comput. Vis. Pattern Recognit.}, 2025, pp. 10510--10522.

\bibitem{arampatzakis2024monocular}
V. Arampatzakis, G. Pavlidis, N. Mitianoudis, and N. Papamarkos, ``Monocular Depth Estimation: A Thorough Review,'' \textit{IEEE Trans. Pattern Anal. Mach. Intell.}, vol. 46, no. 4, pp. 2396--2414, 2024, doi: 10.1109/TPAMI.2023.3330944.

\bibitem{park2024singleview}
B. Park, H. Go, and C. Kim, ``Bridging Implicit and Explicit Geometric Transformation for Single-Image View Synthesis,'' \textit{IEEE Trans. Pattern Anal. Mach. Intell.}, vol. 46, no. 9, pp. 6326--6340, 2024, doi: 10.1109/TPAMI.2024.3378757.

\bibitem{chen2023scenedreamer}
Z. Chen, G. Wang, and Z. Liu, ``SceneDreamer: Unbounded 3D Scene Generation From 2D Image Collections,'' \textit{IEEE Trans. Pattern Anal. Mach. Intell.}, vol. 45, no. 12, pp. 15562--15576, 2023, doi: 10.1109/TPAMI.2023.3321857.

\bibitem{wang2025vggt}
J. Wang, M. Chen, N. Karaev, A. Vedaldi, C. Rupprecht, and D. Novotny, ``VGGT: Visual Geometry Grounded Transformer,'' in \textit{Proc. IEEE/CVF Conf. Comput. Vis. Pattern Recognit.}, 2025, pp. 5294--5306.

\bibitem{wu2025diorama}
Q. Wu, D. Iliash, D. Ritchie, M. Savva, and A. X. Chang, ``Diorama: Unleashing Zero-shot Single-view 3D Indoor Scene Modeling,'' 2025, arXiv:2411.19492.

\bibitem{yao2025cast} K. Yao, L. Zhang, X. Yan, Y. Zeng, Q. Zhang, L. Xu, W. Yang, J. Gu, and J. Yu, ``CAST: Component-Aligned 3D Scene Reconstruction from an RGB Image,'' ACM Trans. Graph., vol. 44, no. 4, Art. no. 86, 19 pages, Aug. 2025, doi: .

\bibitem{xiang2025trellis}
J. Xiang, Z. Lv, S. Xu, Y. Deng, R. Wang, B. Zhang, D. Chen, X. Tong, and J. Yang, ``Structured 3D Latents for Scalable and Versatile 3D Generation,'' 2025, arXiv:2412.01506.

\bibitem{wu2025amodal3r}
T. Wu, C. Zheng, F. Guan, A. Vedaldi, and T.-J. Cham, ``Amodal3R: Amodal 3D Reconstruction from Occluded 2D Images,'' 2025, arXiv:2503.13439.

\bibitem{seo2024genwarp}
J. Seo, K. Fukuda, T. Shibuya, T. Narihira, N. Murata, S. Hu, C.-H. Lai, S. Kim, and Y. Mitsufuji, ``GenWarp: Single Image to Novel Views with Semantic-Preserving Generative Warping,'' 2024, arXiv:2405.17251.

\bibitem{robbyant2026lingbot}
Robbyant Team, Z. Gao, Q. Wang, Y. Zeng, J. Zhu, K. L. Cheng, Y. Li, H. Wang, Y. Xu, S. Ma, Y. Chen, J. Liu et al., ``Advancing Open-source World Models,'' 2026, arXiv:2601.20540.

\bibitem{wang2024motionctrl}
Z. Wang, Z. Yuan, X. Wang, Y. Li, T. Chen, M. Xia, P. Luo, and Y. Shan, ``MotionCtrl: A Unified and Flexible Motion Controller for Video Generation,'' in \textit{ACM SIGGRAPH 2024 Conference Papers}, 2024, pp. 1--11.

\bibitem{zhang2026symphomotion}
G. Zhang, Y. Chen, X. Xiang, J. Huang, Z. Wang, and L. Jiang, ``SymphoMotion: Joint Control of Camera Motion and Object Dynamics for Coherent Video Generation,'' in \textit{Proceedings of the IEEE/CVF Conference on Computer Vision and Pattern Recognition (CVPR)}, June 2026, pp. 11127--11137.

\bibitem{zhang2025tora}
Z. Zhang, J. Liao, M. Li, Z. Dai, B. Qiu, S. Zhu, L. Qin, and W. Wang, ``Tora: Trajectory-Oriented Diffusion Transformer for Video Generation,'' in \textit{Proc. IEEE/CVF Conf. Computer Vision and Pattern Recognition}, 2025, pp. 2063--2073.

\bibitem{zhao2026minwm}
M. Zhao, H. Zhu, B. Yan, Z. Zhou, Y. Chen, W. Sun, K. Zheng, G. He, X. Yang, C. Li et al., ``minWM: A Full-Stack Open-Source Framework for Real-Time Interactive Video World Models,'' 2026, arXiv:2605.30263.

\bibitem{bansal2025videophy2}
H. Bansal, C. Peng, Y. Bitton, R. Goldenberg, A. Grover, and K.-W. Chang, ``VideoPhy-2: A Challenging Action-Centric Physical Commonsense Evaluation in Video Generation,'' 2025, arXiv:2503.06800.

\bibitem{zhang2026physomni}
X. Zhang, Y. Chen, Y. Fang, W. Qu, H. Huang, C. Zhang, F. Xu, and X. Li, ``PhysOmni: Physics-Grounded Multi-Object Scene Generation from a Single Image with Real-Time Interaction,'' in \textit{Proc. 34th ACM Int. Conf. Multimedia (MM)}, 2026, to appear.

\bibitem{carion2025sam3segmentconcepts}
N. Carion, L. Gustafson, Y.-T. Hu et al., ``SAM 3: Segment Anything with Concepts,'' 2025, arXiv:2511.16719.

\bibitem{chen2025sam}
SAM 3D Team, X. Chen, F.-J. Chu et al., ``SAM 3D: 3Dfy Anything in Images,'' 2025, arXiv:2511.16624.

\bibitem{suvorov2021resolution}
R. Suvorov, E. Logacheva, A. Mashikhin et al., ``Resolution-Robust Large Mask Inpainting with Fourier Convolutions,'' 2021, arXiv:2109.07161.

\bibitem{tpami_vit_survey}
K. Han, Y. Wang, H. Chen, X. Chen, J. Guo, Z. Liu, Y. Tang, A. Xiao, C. Xu, Y. Xu, Z. Yang, Y. Zhang, and D. Tao, ``A Survey on Vision Transformer,'' \textit{IEEE Trans. Pattern Anal. Mach. Intell.}, vol. 45, no. 1, pp. 87--110, 2023, doi: 10.1109/TPAMI.2022.3152247.

\bibitem{tpami_foundation_models}
M. Awais, M. Naseer, S. Khan, R. M. Anwer, H. Cholakkal, M. Shah, M.-H. Yang, and F. S. Khan, ``Foundation Models Defining a New Era in Vision: A Survey and Outlook,'' \textit{IEEE Trans. Pattern Anal. Mach. Intell.}, vol. 47, no. 4, pp. 2245--2264, 2025, doi: 10.1109/TPAMI.2024.3506283.

\bibitem{tpami_video_prediction_review}
S. Oprea, P. Martinez-Gonzalez, A. Garcia-Garcia, J. A. Castro-Vargas, S. Orts-Escolano, J. Garcia-Rodriguez, and A. Argyros, ``A Review on Deep Learning Techniques for Video Prediction,'' \textit{IEEE Trans. Pattern Anal. Mach. Intell.}, vol. 44, no. 6, pp. 2806--2826, 2022, doi: 10.1109/TPAMI.2020.3045007.

\bibitem{tpami_memcnet}
W. Bao, W.-S. Lai, X. Zhang, Z. Gao, and M.-H. Yang, ``MEMC-Net: Motion Estimation and Motion Compensation Driven Neural Network for Video Interpolation and Enhancement,'' \textit{IEEE Trans. Pattern Anal. Mach. Intell.}, vol. 43, no. 3, pp. 933--948, 2021, doi: 10.1109/TPAMI.2019.2941941.

\bibitem{tpami_video_inpainting}
R. Szeto, X. Sun, K. Lu, and J. J. Corso, ``A Temporally-Aware Interpolation Network for Video Frame Inpainting,'' \textit{IEEE Trans. Pattern Anal. Mach. Intell.}, vol. 42, no. 5, pp. 1053--1068, 2020, doi: 10.1109/TPAMI.2019.2951667.

\bibitem{tpami_vos_memory}
H. Seong, J. Hyun, and E. Kim, ``Video Object Segmentation Using Kernelized Memory Network With Multiple Kernels,'' \textit{IEEE Trans. Pattern Anal. Mach. Intell.}, vol. 45, no. 2, pp. 2595--2612, 2023, doi: 10.1109/TPAMI.2022.3163375.

\bibitem{tpami_tracking_survey}
S. Javed, M. Danelljan, F. S. Khan, M. H. Khan, M. Felsberg, and J. Matas, ``Visual Object Tracking With Discriminative Filters and Siamese Networks: A Survey and Outlook,'' \textit{IEEE Trans. Pattern Anal. Mach. Intell.}, vol. 45, no. 5, pp. 6552--6574, 2023, doi: 10.1109/TPAMI.2022.3212594.

\bibitem{tpami_tracking_experimental}
A. W. M. Smeulders, D. M. Chu, R. Cucchiara, S. Calderara, A. Dehghan, and M. Shah, ``Visual Tracking: An Experimental Survey,'' \textit{IEEE Trans. Pattern Anal. Mach. Intell.}, vol. 36, no. 7, pp. 1442--1468, 2014, doi: 10.1109/TPAMI.2013.230.

\bibitem{tpami_pointcloud_survey}
Y. Guo, H. Wang, Q. Hu, H. Liu, L. Liu, and M. Bennamoun, ``Deep Learning for 3D Point Clouds: A Survey,'' \textit{IEEE Trans. Pattern Anal. Mach. Intell.}, vol. 43, no. 12, pp. 4338--4364, 2021, doi: 10.1109/TPAMI.2020.3005434.

\bibitem{tpami_depth_completion}
J. Hu, C. Bao, M. Ozay, C. Fan, Q. Gao, H. Liu, and T. L. Lam, ``Deep Depth Completion from Extremely Sparse Data: A Survey,'' \textit{IEEE Trans. Pattern Anal. Mach. Intell.}, vol. 45, no. 7, pp. 8244--8264, 2023, doi: 10.1109/TPAMI.2022.3229090.

\bibitem{tpami_geonetpp}
X. Qi, Z. Liu, R. Liao, P. H. S. Torr, R. Urtasun, and J. Jia, ``GeoNet++: Iterative Geometric Neural Network with Edge-Aware Refinement for Joint Depth and Surface Normal Estimation,'' \textit{IEEE Trans. Pattern Anal. Mach. Intell.}, vol. 44, no. 2, pp. 969--984, 2022, doi: 10.1109/TPAMI.2020.3020800.

\bibitem{tpami_dense3d}
C. Hane, C. Zach, A. Cohen, and M. Pollefeys, ``Dense Semantic 3D Reconstruction,'' \textit{IEEE Trans. Pattern Anal. Mach. Intell.}, vol. 39, no. 9, pp. 1730--1743, 2017, doi: 10.1109/TPAMI.2016.2613051.

\bibitem{tpami_novel_views}
K. Rematas, C. H. Nguyen, T. Ritschel, M. Fritz, and T. Tuytelaars, ``Novel Views of Objects from a Single Image,'' \textit{IEEE Trans. Pattern Anal. Mach. Intell.}, vol. 39, no. 8, pp. 1576--1590, 2017, doi: 10.1109/TPAMI.2016.2601093.

\bibitem{tpami_nonrigid_registration}
J. Huang, T. Birdal, Z. Gojcic, L. J. Guibas, and S.-M. Hu, ``Multiway Non-Rigid Point Cloud Registration via Learned Functional Map Synchronization,'' \textit{IEEE Trans. Pattern Anal. Mach. Intell.}, vol. 45, no. 2, pp. 2038--2053, 2023, doi: 10.1109/TPAMI.2022.3164653.

\bibitem{tpami_pixel2meshpp}
C. Wen, Y. Zhang, C. Cao, Z. Li, X. Xue, and Y. Fu, ``Pixel2Mesh++: 3D Mesh Generation and Refinement From Multi-View Images,'' \textit{IEEE Trans. Pattern Anal. Mach. Intell.}, vol. 45, no. 2, pp. 2166--2180, 2023, doi: 10.1109/TPAMI.2022.3169735.

\bibitem{tpami_epc}
C. Luo, Z. Yang, P. Wang, Y. Wang, W. Xu, R. Nevatia, and A. Yuille, ``Every Pixel Counts ++: Joint Learning of Geometry and Motion with 3D Holistic Understanding,'' \textit{IEEE Trans. Pattern Anal. Mach. Intell.}, vol. 42, no. 10, pp. 2624--2641, 2020, doi: 10.1109/TPAMI.2019.2930258.

\bibitem{tpami_surRF}
J. Zhang, M. Ji, G. Wang, Z. Xue, S. Wang, and L. Fang, ``SurRF: Unsupervised Multi-View Stereopsis by Learning Surface Radiance Field,'' \textit{IEEE Trans. Pattern Anal. Mach. Intell.}, vol. 44, no. 11, pp. 7912--7927, 2022, doi: 10.1109/TPAMI.2021.3116695.

\bibitem{tpami_refnerf}
D. Verbin, P. Hedman, B. Mildenhall, T. Zickler, J. T. Barron, and P. P. Srinivasan, ``Ref-NeRF: Structured View-Dependent Appearance for Neural Radiance Fields,'' \textit{IEEE Trans. Pattern Anal. Mach. Intell.}, vol. 47, no. 11, pp. 9426--9437, 2025, doi: 10.1109/TPAMI.2024.3360018.

\bibitem{tpami_physics_gan}
J. Pan, J. Dong, Y. Liu, J. Zhang, J. Ren, J. Tang, Y.-W. Tai, and M.-H. Yang, ``Physics-Based Generative Adversarial Models for Image Restoration and Beyond,'' \textit{IEEE Trans. Pattern Anal. Mach. Intell.}, vol. 43, no. 7, pp. 2449--2462, 2021, doi: 10.1109/TPAMI.2020.2969348.

\bibitem{tpami_partialconv}
G. Liu, A. Dundar, K. J. Shih, T.-C. Wang, F. A. Reda, K. Sapra, Z. Yu, X. Yang, A. Tao, and B. Catanzaro, ``Partial Convolution for Padding, Inpainting, and Image Synthesis,'' \textit{IEEE Trans. Pattern Anal. Mach. Intell.}, vol. 45, no. 5, pp. 6096--6110, 2023, doi: 10.1109/TPAMI.2022.3209702.

\bibitem{Qwen2VL}
P. Wang, S. Bai, S. Tan et al., ``Qwen2-VL: Enhancing Vision-Language Model's Perception of the World at Any Resolution,'' 2024, arXiv:2409.12191.

\bibitem{Genesis}
Genesis Authors, ``Genesis: A Generative and Universal Physics Engine for Robotics and Beyond,'' 2024. [Online]. Available: \url{https://github.com/Genesis-Embodied-AI/Genesis}

\bibitem{featherstone2014rigid}
R. Featherstone, \textit{Rigid Body Dynamics Algorithms}. New York, NY, USA: Springer, 2008.


\bibitem{an2025aiflowperspectivesscenarios}
H. An et al., ``AI Flow: Perspectives, Scenarios, and Approaches,'' 2025, arXiv:2506.12479. [Online]. Available: \url{https://arxiv.org/abs/2506.12479}.

\bibitem{shao2025aiflownetworkedge}
J. Shao and X. Li, ``AI Flow at the Network Edge,'' 2025, arXiv:2411.12469. [Online]. Available: \url{https://arxiv.org/abs/2411.12469}.

\bibitem{chen2025teleworlddynamicmultimodalsynthesis}
Y. Chen et al., ``TeleWorld: Towards Dynamic Multimodal Synthesis with a 4D World Model,'' 2025, arXiv:2601.00051. [Online]. Available: \url{https://arxiv.org/abs/2601.00051}.

\bibitem{chen2024cascadezero123imagehighlyconsistent}
Y. Chen et al., ``Cascade-Zero123: One Image to Highly Consistent 3D with Self-Prompted Nearby Views,'' 2024, arXiv:2312.04424. [Online]. Available: \url{https://arxiv.org/abs/2312.04424}.

\bibitem{chen2024liftimage3dliftingsingleimage}
Y. Chen et al., ``LiftImage3D: Lifting Any Single Image to 3D Gaussians with Video Generation Priors,'' 2024, arXiv:2412.09597. [Online]. Available: \url{https://arxiv.org/abs/2412.09597}.

\bibitem{xiang2025macrofrommicroplanninghighqualityparallelized}
X. Xiang et al., ``Macro-from-Micro Planning for High-Quality and Parallelized Autoregressive Long Video Generation,'' 2025, arXiv:2508.03334. [Online]. Available: \url{https://arxiv.org/abs/2508.03334}.

\bibitem{xiang2026pathwisetesttimecorrectionautoregressive}
X. Xiang et al., ``Pathwise Test-Time Correction for Autoregressive Long Video Generation,'' 2026, arXiv:2602.05871. [Online]. Available: \url{https://arxiv.org/abs/2602.05871}.

\bibitem{wang2026directingworldfastautoregressive}
H. Wang, Y. Chen, H. Huang, C. Zhang, and X. Li, ``Directing the World: Fast Autoregressive Video Generation with Compositional Human-Camera Control,'' 2026, arXiv:2606.27964. [Online]. Available: \url{https://arxiv.org/abs/2606.27964}.

\bibitem{xiang2025makeefficientdynamicsparse}
X. Xiang and Q. Fan, ``Make It Efficient: Dynamic Sparse Attention for Autoregressive Image Generation,'' 2025, arXiv:2506.18226. [Online]. Available: \url{https://arxiv.org/abs/2506.18226}.

\bibitem{huang2025zero}
Y. Huang, Y. Chen, L. Ding, X. Zhang, W. Dai, J. Zou, H. Xiong, and Q. Tian, ``IM-Zero: Instance-Level Motion Controllable Video Generation in a Zero-Shot Manner,'' in \textit{Proc. IEEE/CVF Conf. Comput. Vis. Pattern Recognit.}, 2025, pp. 7265--7275.

\bibitem{huang2024domainfusion}
Y. Huang, Y. Chen, Y. Liu, X. Zhang, W. Dai, H. Xiong, and Q. Tian, ``DomainFusion: Generalizing to Unseen Domains with Latent Diffusion Models,'' in \textit{Proc. Eur. Conf. Comput. Vis.}, 2024, pp. 480--498.

\bibitem{wen2025metricsolverslidinganchoredmetric}
T. Wen, J. Wang, Y. Chen, S. Xu, C. Zhang, and X. Li, ``Metric-Solver: Sliding Anchored Metric Depth Estimation from a Single Image,'' 2025, arXiv:2504.12103. [Online]. Available: \url{https://arxiv.org/abs/2504.12103}.

\bibitem{chen2026full4dgeneratingfullscope4d}
T. Chen, K. Hao, Y. Chen, Z. Cheng, R. Xie, L. Song, H. Huang, C. Zhang, and X. Li, ``Full-4D: Generating Full-Scope 4D Scenes from a Single-View Video,'' 2026, arXiv:2605.25500. [Online]. Available: \url{https://arxiv.org/abs/2605.25500}.

\end{thebibliography}
\end{document}